%% file: main.tex
\documentclass[11pt]{article}
\usepackage[preprint]{acl}

\usepackage{times}
\usepackage{latexsym}
\usepackage{enumitem}
\usepackage{amsmath}
\usepackage{amssymb}
\usepackage{tabularx}
\usepackage[table]{xcolor}
\definecolor{lightblue}{RGB}{220,235,250}

\usepackage[T1]{fontenc}

\usepackage[utf8]{inputenc}
\usepackage{multirow}
\usepackage{adjustbox}
\usepackage{hyperref}
\usepackage{float}
\usepackage{siunitx}
\usepackage{cuted}
\usepackage{caption}
\usepackage{booktabs}
\usepackage{placeins}
\usepackage{xspace}
\newcommand{\ie}{i.e.\xspace}

\usepackage{microtype}

\usepackage{inconsolata}

\usepackage{graphicx}
\usepackage{booktabs}
\usepackage{multirow}
\usepackage{multicol}
\usepackage{subcaption}
\usepackage{amsmath}
\usepackage[most,skins,theorems]{tcolorbox}
\usepackage{xcolor}

\definecolor{lightpink}{RGB}{238,206,205}
\definecolor{grayhighlight}{RGB}{240,240,240}
\definecolor{lightblue}{RGB}{210, 222, 238}

\tcbset{
  aibox/.style={
    width=\linewidth,
    top=8pt,
    bottom=4pt,
    colback=lightpink,
    colframe=black,
    colbacktitle=black,
    enhanced,
    center,
    attach boxed title to top left={yshift=-0.1in,xshift=0.15in},
    boxed title style={boxrule=0pt,colframe=white,},
  }
}

\newtcolorbox{AIbox}[2][]{aibox,title=#2,#1}
\title{Revisiting the Uniform Information Density Hypothesis in LLM Reasoning}

\author{
  Minju Gwak$^{1,2}$ \quad
  Guijin Son$^{2}$ \quad
  Jaehyung Kim$^{1}$ \\
  $^{1}$Yonsei University \quad
  $^{2}$OneLine AI \\
  \texttt{mjgwak@yonsei.ac.kr, jaehyungk@yonsei.ac.kr}
}

\begin{document}
\maketitle
\begin{abstract}
The Uniform Information Density (UID) hypothesis proposes that effective communication is achieved by maintaining a stable flow of information. In this work, we revisit this principle in the context of Large Language Model (LLM) reasoning, asking whether step-level uniformity reflects reasoning quality. To this end, we introduce a novel framework to quantify uniformity of information flow at both local and global levels, using an entropy-based stepwise density metric. Across experiments on seven reasoning benchmarks, we see a counter-intuitive pattern: while high-quality reasoning exhibit smooth step-by-step transitions (\textit{local uniformity}) and structured, non-uniform information flow at the trajectory level (\textit{global non-uniformity}). The results demonstrate that these uniformities outperform alternative internal signals as predictors of reasoning quality, and such divergence with human communication is not a model deficiency, but a byproduct of distinct objectives between human communication and LLM reasoning.
\footnote{Code is released at: \url{https://github.com/talzoomanzoo/uid-reasoning}}
\end{abstract}

\input{1_intro}
\input{3_method}
\input{5_results}
\input{2_related_work}
\input{6_conclusion}
\bibliography{custom}
\appendix
\input{7_appendix}
\end{document}

%% file: 1_intro.tex
\section{Introduction}

Chain-of-Thought (CoT) reasoning has become a central technique for enhancing large language models (LLMs) on complex reasoning tasks
\citep{wei2023chainofthoughtpromptingelicitsreasoning, kojima2023largelanguagemodelszeroshot, chae2023dialoguechainofthoughtdistillationcommonsenseaware}. 
By generating step-by-step rationales, CoT enables models to decompose problems into simpler subproblems and thereby improve accuracy \citep{golovneva2023roscoesuitemetricsscoring, prasad2023recevalevaluatingreasoningchains, yao2023treethoughtsdeliberateproblem}. 
Despite these successes, recent studies have highlighted the fragility of this approach \citep{zhao2025chainofthoughtreasoningllmsmirage}.
For example, the intermediate rationales are often logically inconsistent or incoherent, and hence models fail to generalize out-of domain tasks even when producing lengthy reasoning traces \citep{shojaee2025illusionthinkingunderstandingstrengths}.  
This raises a critical question: \textit{how can we determine whether LLMs are reasoning effectively, rather than merely generating superficially coherent text?}

\begin{figure}[t]
    \centering
    \begin{subfigure}[t]{\linewidth}
        \centering
        \includegraphics[width=\linewidth]{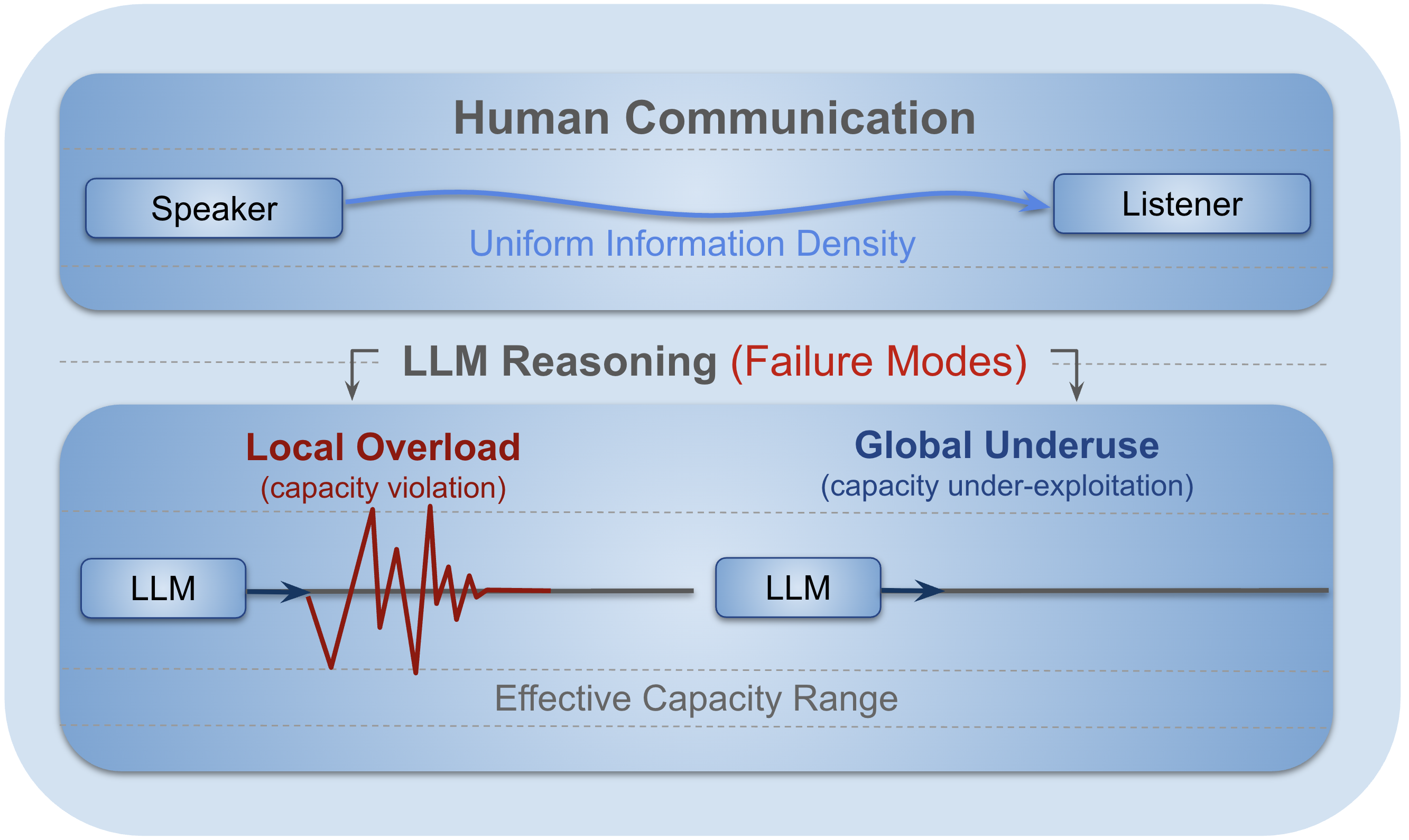}
        \label{fig:correct_averaged}
    \end{subfigure}
    
    \caption{\textbf{Reasoning as information flow.} Human communication distributes information smoothly to respect channel capacity, enabling successful understanding. LLM reasoning transmits information across reasoning steps; failures arise from local overload (sharp spikes) and underuse (flat trivial trajectory).
}
\vspace{-0.2in}
    \label{fig:figure1}
    \label{fig:correct-incorrect}
\end{figure}
\label{fig:reasoning_channel}

Clues may lie in human communication itself; the psycholinguistic hypothesis of Uniform Information Density (UID) proposes that speakers distribute information as evenly as possible to balance clarity and efficiency \citep{fenk1980konstanz, genzel2002entropy, clark2023crosslinguisticpressureuniforminformation,jaeger2006speakersoptimizeinfo}. 
Namely, a relatively uniform flow of information is necessary for effective communication \citep{meister2021revisitinguniforminformationdensity,aylett2004smooth}, aligned with the limits of human cognitive processing. 
When this balance is disrupted by too much or too little information, communication deteriorates. 
Motivated by this, we ask \textit{whether a similar principle governs reasoning in LLMs.} 
As human speakers maintain balanced information flow to support comprehension, effective reasoning traces may require comparable uniformity across steps.  
Recent findings in cognitive science support this view: \citet{bhambri2025cognitivelyinterpretablereasoningtraces} shows that reasoning paths interpretable to humans are also easier for models to generate and learn, suggesting a shared structure between human cognition and machine reasoning. 

To investigate this, we focus on analyzing the information flow of LLM-generated reasoning traces on challenging mathematical benchmarks. 
Specifically, we begin by defining per-step measurements of information density using entropy of predictive distribution, and examine their relationship to answer correctness. 
We then introduce two complementary metrics to quantify uniformity at both global and local levels. 
Our experiments reveal a counter-intuitive pattern: unlike human communication, successful LLM reasoning exhibits high local uniformity but low global uniformity. 
In our experiments across seven challenging reasoning benchmarks and three LLMs, these uniformity consistently outperform conventional approaches in identifying high-quality trajectories for Best-of-N sampling.
We find that this divergence is not a model deficiency, but rather an instrumental byproduct of the distinct objectives between human communication and LLM reasoning.

Overall, our contributions are threefold:
\begin{itemize}[leftmargin=5.5mm,topsep=0pt]
    \vspace{0.05in}
    \item[$\circ$] To our knowledge, we are the first to revisit the Uniform Information Density (UID) hypothesis in the context of LLM reasoning. \vspace{-0.1in}
    \item[$\circ$] Contrary to our hypothesis, we find that reasoning patterns characterized by global non-uniformity and local uniformity in surprisal correlate with reasoning success on challenging mathematical reasoning tasks. \vspace{-0.1in}
    \item[$\circ$] Extensive analyses show that deviations from such patterns serve as a trace-level internal signal for predicting failure cases, enabling complementary improvements to response-level aggregation and LLM reasoning evaluation.
\end{itemize}

%% file: 3_method.tex
\section{Exploring the Uniform Information Density Hypothesis in LLM Reasoning}
\subsection{Background: the UID hypothesis}
The UID hypothesis considers language as a signal transmitted through a noisy channel with limited capacity \citep{meister2021revisitinguniforminformationdensity, tsipidi2024surpriseuniforminformationdensity}. 
Then, UID posits that speakers aim to convey information efficiently without overwhelming the listener's processing resources. 
Formally, let an utterance $\textbf{u}$ = $[u_1$,$u_2$, \dots, $u_N]$ be a sequence of $N$ linguistic units, such as words, subwords, or characters, depending on the granularity of representation. 
For each unit $u_n$, \textit{surprisal} is defined as its unexpectedness, given its previous context:
\begin{equation*}
    s(u_n) = -\log P(u_n \mid \textbf{u}_{<n}),
\end{equation*}
where $P(u_n |\textbf{u}_{<n})$ is the probability of seeing unit utterance $u_n$ after the earlier sequence $\textbf{u}_{<n}=[u_1,\dots,u_{n-1}]$. 
Intuitively, the unit with high surprisal is very unexpected and therefore hard to process for the listener. 
Then, the total processing effort for the utterance $\textbf{u}$ is expressed as:
\begin{equation*}
    E_{\tt process}(\textbf{u}) \propto \sum_{n=1}^{N} s(u_n)^{k} + c \cdot N.
\end{equation*}
for some constant $c > 0$ and $k > 1$. 
Here, the exponent $k$ encodes the super-linear nature of processing effort, where rare or unexpected units impose larger effort than predictable ones.  
Under this formulation, uniform surprisals across units minimize total processing effort, whereas “spiky” linguistic signal with highly uneven surprisal values increase the burden of communication~\citep{meister2021revisitinguniforminformationdensity}.

While UID has been validated in human language, its implications for machine reasoning remain underexplored. 
LLMs, or more specifically, recent reasoning models such as Deepseek-R1 \citep{deepseekai2025deepseekr1incentivizingreasoningcapability} and Qwen3 \citep{yang2025qwen3technicalreport} generate step-by-step CoT traces, similar to how human speech unfold over time. 
If we treat each reasoning step $z_i$ like a unit with surprisal $s(z_i)$, a single reasoning trace $\textbf{z}$ = $[z_1$, $z_2$, \dots, $z_N]$ can be analyzed in the same way to have the total reasoning effort as below:
\begin{equation*}
    E_{\tt reason}(\textbf{z}) \propto \sum_{n=1}^{N} s(z_n)^{k} + c \cdot N.
\end{equation*}
Then, a natural question arises: \textit{does the UID hypothesis hold for good reasoning patterns in LLMs?} 
A smooth, uniform surprisal profile may reflect clear and logical reasoning, while sharp spikes may signal confusion or errors. 
Therefore, in this work, we validate UID hypothesis beyond psycholinguistics to CoT reasoning of LLMs, offering a new lens on why reasoning models succeed or fail.

\subsection{Preliminary analyses: Step-wise information density in CoTs of LLMs}

We start by defining the step-level information density $ID_i$ for a reasoning trace $\mathbf{z}=[z_1,\dots,z_N]$, where each reasoning step $z_i$ is composed of $M_i$ tokens, \textit{i.e.}, $z_i=[x_1,\dots,x_{M_i}]$. 
We divide the given reasoning trace into multiple reasoning steps using \texttt{\textbackslash n\textbackslash n}, following \citet{lightman2023letsverifystepstep}.\footnote{While we adopt newline-based segmentation, we demonstrate that our findings are robust to alternative stepwise segmentation strategies in Appendix~\ref{app:segmentation_robustness}.} 
Let $p_t$ be the predictive distribution over the vocabulary $\mathcal{V}$ at the token position $t$.
Then, to characterize $ID_i$, we consider entropy over tokens in each step:

\begin{equation*}
    H_t = - \sum_{v \in \mathcal{V}} p_t(v) \log p_t(v),
\end{equation*}
and step-level information density with entropy is:
\begin{equation*}
    ID_i = \frac{1}{M_i} \sum_{t=1}^{M_i} H_t.
\end{equation*}
\begin{figure}[t]
    \centering
    \begin{subfigure}[t]{\linewidth}
        \centering
        \includegraphics[width=1.0\linewidth]{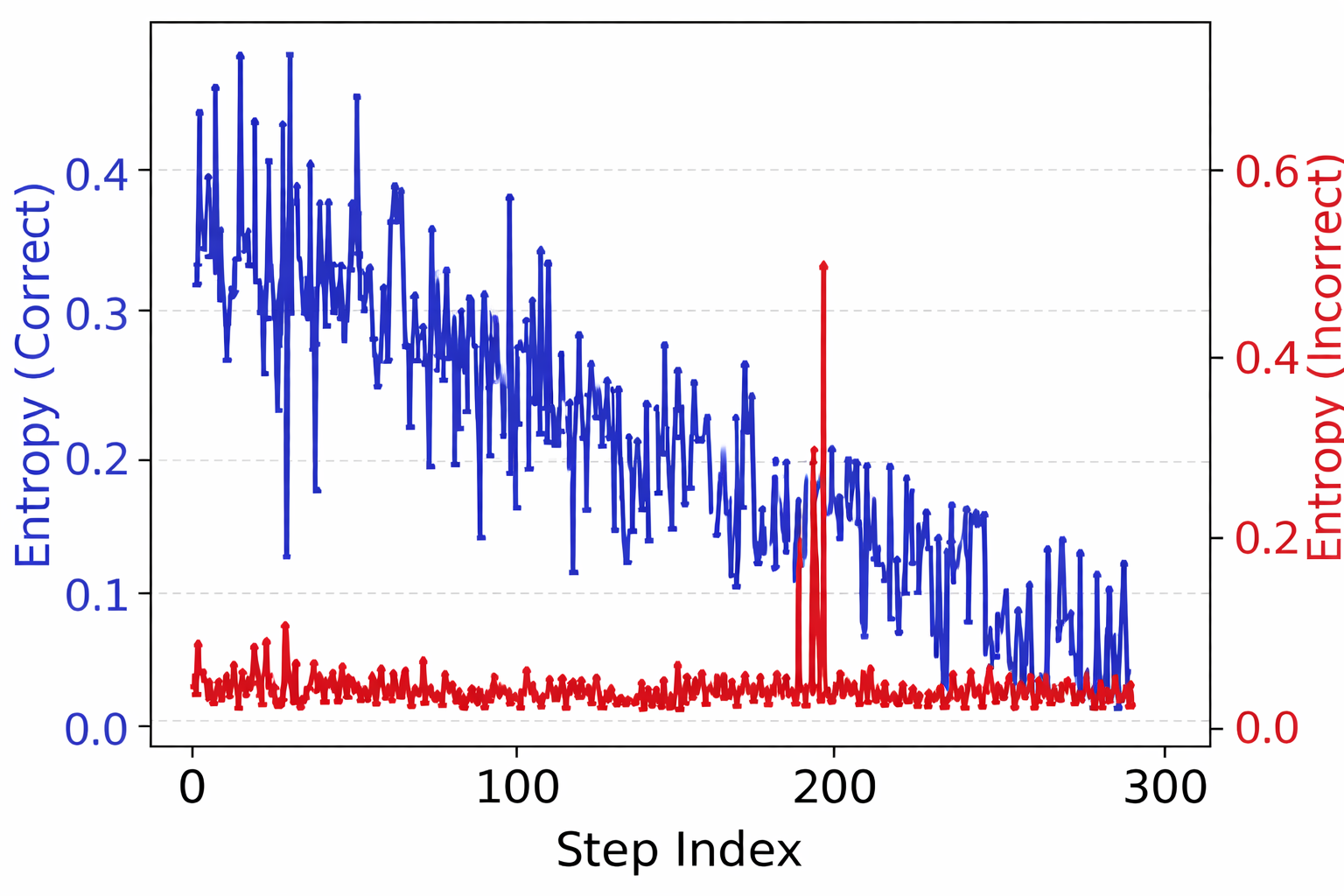}
        \label{fig:correct_averaged}
    \end{subfigure}
    
    \caption{Averaged $ID_i$ scores of LLM reasoning traces on AIME2025. Correct traces show \textbf{a downward trend with smooth decay}, while incorrect traces show noisy entropy with \textbf{unresolved spikes}.}
    \label{fig:figure1}
    \label{fig:correct-incorrect}
\end{figure}
\paragraph{Justifications for using entropy as a proxy.}We use entropy as a proxy for information density because it reflects both model confidence and variability in reasoning;  
low entropy indicates confident predictions while higher entropy implies uncertainty between multiple plausible continuations~\citep{shannon1948mathematicaltheorycommunication, kuhn2013applied}. 
Also, in an information-theoretic perspective, entropy quantifies the expected number of bits required to encode the predictive distribution, where higher values corresponding to richer informational content~\citep{cover2006elements}. 
Therefore, aggregating entropy across tokens offers a compact and interpretable signal of reasoning difficulty.
Furthermore, our experiments (see Appendix~\ref{app:empirical}) suggest that using entropy as information density is more effective, compared to other candidates such as log-probability and confidence-based methods.

Figure~\ref{fig:correct-incorrect} compares the evolution of $ID_i$ between the averaged reasoning traces for correct and incorrect solutions in AIME2025, respectively. 
Here, correct traces exhibit a clear global trend: entropy begins with exploratory fluctuations, stabilizes in mid-trace, and then steadily decays toward near zero, reflecting a structured process of resolution and convergence on the final answer. 
In contrast, incorrect traces instead displays a flat, noisy entropy trajectory with occasional sharp spikes.

\paragraph{Motivation for a structural perspective.} As we examine entropy peaks at the level of individual reasoning traces, we find that interpreting them solely through the semantic content of the text is difficult (see Appendix ~\ref{app:trace}). 
To be specific, (1) entropy levels in both correct and incorrect traces can be quite varying, (2) number of transition words \textit{(\ie, But, Alternatively, Wait)} may appear more at correct traces contrary to our intuition, and (3) wrong reasoning traces may be concise and have fewer number of steps. 
More importantly, such an approach does not provide a consistent basis for interpretation across various benchmarks, highlighting the need for a unified structural perspective. 
Building on such motivation, we introduce a framework for measuring the uniformity of information density in reasoning traces.

\subsection{Measuring global and local uniformity of information density in CoTs of LLMs}

To quantify the uniformity of information density in a reasoning trace, we first distinguish between two complementary notions of uniformity that have been discussed in prior psycholinguistic work~\citep{meister2021revisitinguniforminformationdensity, collins2014information}. 
\textbf{Global uniformity} characterizes whether information is distributed evenly across the entire trace, corresponding to a relatively stable surprisal level over long horizons.
In contrast, \textbf{local uniformity} captures whether information changes smoothly between adjacent steps, reflecting gradual and coherent transitions rather than abrupt jumps. 
These two notions capture uniformity in different ways and therefore diverge substantially in reasoning traces: a trace may appear globally uniform yet contain sharp local disruptions, or conversely, exhibit smooth local transitions while concentrating information unevenly across steps.

Motivated by this distinction, we introduce two complementary UID-based metrics that separately operationalize global and local uniformity in LLM reasoning traces: (1) \emph{global variance} and (2) \emph{local step-to-step spikes and falls}.

\paragraph{Global uniformity via variance.}
Global uniformity is measured by the variance of step-level information density across the entire reasoning trace, capturing \textit{whether the information is evenly distributed or concentrated in a small number of steps.} 
Formally, for a reasoning trace $\mathbf{z}=[z_1,\dots,z_N]$, let us define the non-negative information density vector $\mathbf{u} = [ID_1, \dots, ID_N]$. 
After min--max normalization, where $ID_i' = \frac{ID_i - m}{M - m}$ with $m = \min_{1 \le i \le N} ID_i$ and $M = \max_{1 \le i \le N} ID_i$, we obtain the normalized vector $\tilde{\mathbf{u}} = [ID_1', \dots, ID_N']$. 
The variance of the normalized information density values is then defined as:
\begin{equation}
    \mathrm{Var}(\tilde{\mathbf{u}}) =\frac{1}{N} \sum_{i=1}^{N} \left( ID_i' - \mu \right)^2,
\end{equation}
where $\mu = \frac{1}{N} \sum_{i=1}^{N} ID_i'$.
High variance indicates global non-uniformity, where information is unevenly concentrated across steps, while lower variance corresponds to globally uniform traces.

\paragraph{Local uniformity via step-to-step smoothness.}
Local uniformity captures how smoothly information density evolves between adjacent reasoning steps, measuring whether uncertainty is resolved gradually or through abrupt transitions. 
Given $\tilde{\mathbf{u}}$, we define the step-to-step change as $\Delta_i = ID_i' - ID_{i-1}'$ for $i = 2, \dots, N$, and compute the mean and standard deviation of the change sequence as $\mu_{\Delta} = \frac{1}{N-1} \sum_{i=2}^{N} \Delta_i$ and $\sigma_{\Delta} = \sqrt{\frac{1}{N-1} \sum_{i=2}^{N} (\Delta_i - \mu_{\Delta})^2}$. 
To identify significant local disruptions, we define thresholds $T^{+} = \mu_{\Delta} + \tau\sigma_{\Delta}$ and $T^{-} = \mu_{\Delta} - \tau\sigma_{\Delta}$, where $\tau \in \{2,3\}$.
Then, an \emph{upward spike} is identified when $\Delta_i > T^{+}$, and a \emph{downward fall} when $\Delta_i < T^{-}$. 
Finally, total local irregularity count is:
\begin{equation}
\begin{aligned}
S_{\tt up}(\tilde{\mathbf{u}})
  &= \sum_{i=2}^{N} \mathbb{I}[\Delta_i > T^{+}], \\
S_{\tt down}(\tilde{\mathbf{u}})
  &= \sum_{i=2}^{N} \mathbb{I}[\Delta_i < T^{-}], \\
S_{\tt local}(\tilde{\mathbf{u}})
  &= S_{\tt up}(\tilde{\mathbf{u}}) + S_{\tt down}(\tilde{\mathbf{u}}).
\end{aligned}
\end{equation}

Intuitively, smaller $S_{\tt local}$ indicates higher local uniformity, therefore reflecting smoother stepwise information flow.

These two metrics, $\mathrm{Var}(\tilde{\mathbf{u}})$ and $S_{\tt local}(\tilde{\mathbf{u}})$, provide complementary trace-level characterization of how information is distributed and evolved during reasoning. 
As reflected in our empirical observations (Figure~\ref{fig:correct-incorrect}), correct reasoning traces tend to exhibit \emph{low local disruption} (\textit{i.e.}, high local uniformity) alongside \emph{structured global non-uniformity} (\textit{i.e.}, low global uniformity), whereas incorrect traces often exhibit large, abrupt fluctuations in local uniformity or consistently high and flat global uniformity that fails to convey informative structure.
This complementary view enables a more precise diagnosis of reasoning quality beyond token-wise confidence or entropy measures.

%% file: 5_results.tex
\input{Tables/MainResults_ds}

\section{Does the UID Hypothesis Hold in LLM Reasoning Traces?}
\paragraph{Setups.}
For the experiments, we use three widely adopted open-source reasoning models, Deepseek-R1-Distill-Qwen-7B \citep{deepseekai2025deepseekr1incentivizingreasoningcapability}, Deepseek-R1-Distill-Llama-8B \citep{deepseekai2025deepseekr1incentivizingreasoningcapability}, and Qwen3-8B \citep{yang2025qwen3technicalreport}.
Also, we use four challenging mathematical benchmarks: AIME2025, BRUMO2025, HMMT2025, and MinervaMath (MM). 
To demonstrate the effectiveness as the selection criteria of Best-of-N, we sample five reasoning traces  for each question, with temperature as 0.6, top-p as 0.95, and top-k as 20. 
Then, we calculate their corresponding UID scores, and select the traces with the highest and lowest scores. 
We consider the following three baselines that leverage internal signals to find a good logical trail: (1) Self-Certainty \citep{kang2025scalablebestofnselectionlarge}, (2) High confidence, and (3) Low entropy. 
Although we evaluate both $\tau$ = 2, 3, we report only the results of $\tau$ = 3 results in the main text. 
Details are presented in Appendix \ref{app:impl}.

\paragraph{Main results.}
Table~\ref{tab:mainresults_ds} shows that UID-guided selection outperforms non-UID baselines across all benchmarks, which indicates that, \textit{unlike human communication postulated by the original UID hypothesis, reasoning success in LLMs is best predicted by a combination of local uniformity and global non-uniformity in information distribution.}

For Deepseek-R1-Distill-Qwen-7B under \emph{local uniformity}, the accuracies are improved by +33\% on AIME2025, +4\% on BRUMO2025, +25\% on HMMT2025, and +3\% on MinervaMath. 
These gains are consistent, indicating that smoothing step-wise information flow reliably benefits reasoning regardless of task domain.

On the other hand, \emph{global non-uniformity} also strengthens performance, particularly on harder benchmarks. 
Compared to Self-Certainty selection, global non-uniformity yields an additional +23\% improvement on BRUMO2025 and +8\% on AIME2025, while maintaining comparable performance on MinervaMath and only marginal degradation on HMMT2025. 
Overall, this setting achieves the strongest average performance, with relative gains reaching up to 32\% over the vanilla accuracy baseline.
Together, these results suggest that \emph{local smoothness and global heterogeneity provides the most stable and informative criterion for selecting high-quality reasoning traces.}

Similar trends are observed for Deepseek-R1-Distill-Llama-8B. 
Under local uniformity, accuracy improves by approximately 15--20\% on AIME2025 and BRUMO2025 relative to the overall-accuracy baseline, while yielding more modest gains 5--10\% on HMMT2025 and MinervaMath.
Enabling global non-uniformity further enhances performance on AIME2025 and HMMT2025, delivering an additional 10--15\% improvement over Self-Certainty selection, while remaining competitive on BRUMO2025 and MinervaMath. 
These results indicate that UID-guided selection remains effective even when the backbone model exhibits weaker absolute reasoning performance, though the gains are naturally bounded by model capacity.

Qwen3-8B follows the same qualitative trend, though with reduced margins. 
Because its baseline accuracy is already strong, UID-guided selection yields more modest gains (approximately 2--5\% on AIME2025 and BRUMO2025), while still consistently outperforming entropy- and confidence-based heuristics. 
This confirms that UID metrics generalize across models but naturally offer larger benefits when baseline reasoning quality is weaker.

\begin{AIbox}{Takeaway for UID-Guided Selection}
Reasoning traces exhibiting \emph{local uniformity} and \emph{global non-uniformity} achieve the highest accuracy, consistently outperforming confidence- and entropy-based selection across benchmarks.
\end{AIbox}

\section{Further Analyses}
\subsection{Trends across different model sizes}

\input{Tables/ScaleModel}
Table~\ref{tab:scalemodel} presents the results with \texttt{Qwen3} models of different sizes (1.7B, 4B, 8B) and reveals clear scaling trends.
As expected, larger models achieve stronger baselines: mean accuracy increases by approximately 86\% from 1.7B to 4B, with a more modest 3\% relative gain from 4B to 8B.
Self-Certainty shows a similar pattern, rising by 62\% from 1.7B to 4B , followed by a 14\% relative decrease at 8B, suggesting diminishing returns and possible overconfidence effects at larger scales.

Despite reduced headroom from stronger baselines, UID-based methods continue to provide meaningful relative improvements.
Local uniformity yields consistent gains across all scales, improving performance by 17\% for 1.7B, 6\% for 4B, and 3\% for 8B, indicating that enforcing local smoothness stabilizes reasoning regardless of model size.
In contrast, global non-uniformity exhibits a pronounced scaling effect: while the 1.7B model improves by only 6\%, the 4B and 8B models gain 2\% and 4\%, respectively, with the 8B model achieving the strongest overall performance in the table.
\textit{These results suggest a size-dependent effect: smaller models benefit more from local smoothing, whereas larger models increasingly exploit global non-uniformity in surprisal distribution.}

\begin{AIbox}{Takeaway 4.1 for Scaling with Model Size}
UID-guided reasoning selection scales with model capacity.
\end{AIbox}

\subsection{Generalization beyond math reasoning}
\input{Tables/NonMath}

To evaluate generalization beyond mathematical reasoning, we examine reasoning performance on non-math benchmarks (GPQA-Diamond (GPQA-D), LSAT-AR, and LSAT-LR).

Table~\ref{tab:scalesample} reports results under different sampling scales. 
We observe a consistent tendency for both DS-R1-Distill-Llama-8B and DS-R1-Distill-Qwen-7B; UID-based strategies consistently improve over the mean-accuracy baseline. 
For DS-R1-Distill-Llama-8B on GPQA-D, locally uniform UID increases accuracy from 0.48 to 0.52, corresponding to an absolute gain of +4\%p compared to mean accuracy. 
Similar trends hold on LSAT tasks: on LSAT-AR, locally uniform UID improves performance by about +7\% relative compared to mean accuracy, while on LSAT-LR, it achieves the best overall score with a +1\%p absolute gain.

For DS-R1-Distill-Qwen-7B, the effect is even more pronounced. 
On LSAT-AR, both locally uniform and globally non-uniform UID reach 0.62 accuracy, improving over the mean baseline (55\%) by +7\%p, or roughly +12.7\% relative improvement. 
On GPQA-D, locally uniform UID matches the strongest non-UID heuristic (self-certainty), while maintaining a +3–4\% relative gain over mean accuracy. 
On LSAT-LR, globally non-uniform UID achieves the best score, yielding a +6\% relative improvement compared to the baseline.

Such results show that UID metrics capture meaningful structural signals even in non-mathematical reasoning. 
Rather than merely matching strong confidence-based heuristics, UID-based selection often yields \emph{consistent percentage-level gains} across diverse reasoning benchmarks.

\begin{AIbox}{Takeaway 4.2 for Domain Generalization}
UID-guided selection yields consistent relative accuracy gains (5--13\%) across non-math reasoning tasks, demonstrating robust generalization beyond math domains.
\end{AIbox}

\input{Tables/ScaleSample}

\subsection{Trends across different sample sizes}

Table~\ref{tab:nomath} reports results under different sampling scales. Across all regimes, we observe consistently higher-performing reasoning traces exhibit \emph{local uniformity} while remaining \emph{globally non-uniform}.

This pattern already emerges under the smallest setting, Sample by 3. Locally uniform traces achieve 0.73 accuracy, outperforming locally non-uniform variants by a clear margin. At the global level, non-uniform traces reach 0.70, while globally uniform traces do not provide additional gains.

When sampled by 5, locally uniform traces achieve 0.69 accuracy, again outperforming locally non-uniform variants. At the global level, non-uniform traces reach 0.70, exceeding globally uniform ones by approximately 6\% relative.

This trend becomes more pronounced as we scale up to Sample by 10. Locally uniform traces further improve to 0.72 accuracy, while globally uniform traces degrade to 0.63, corresponding to a $\sim$12\% relative decrease. Across all sampling settings, locally uniform traces consistently outperform locally non-uniform ones by 3–6 percentage points, while global non-uniformity remains a key factor for maintaining high accuracy.

\begin{AIbox}{Takeaway 4.3 for Scaling with Sample Size}
Effective reasoning benefits from local uniformity but global non-uniformity: stable step-level information flow with trajectory-level variation yields the highest accuracy, even under small sampling budgets.
\end{AIbox}

\begin{figure}[t]
    \centering

    \begin{subfigure}[t]{0.935\linewidth}
        \centering
        \includegraphics[width=0.82\linewidth]{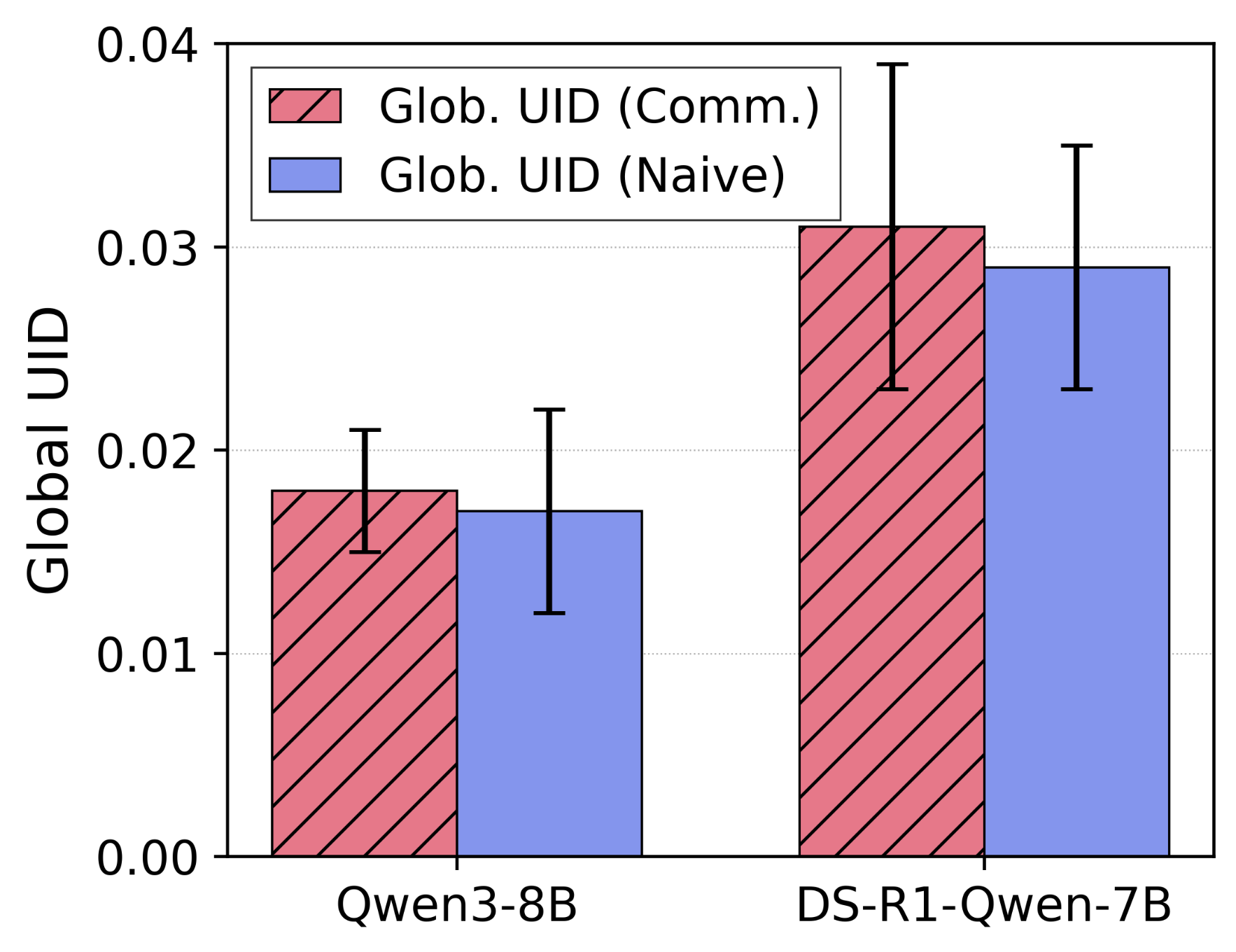}
        \caption{Global UID of comm vs.\ naive prompting}
        \label{fig:uid_global}
    \end{subfigure}

    \vspace{0.5em} 

    \begin{subfigure}[t]{0.94\linewidth}
        \centering
        \includegraphics[width=0.80\linewidth]{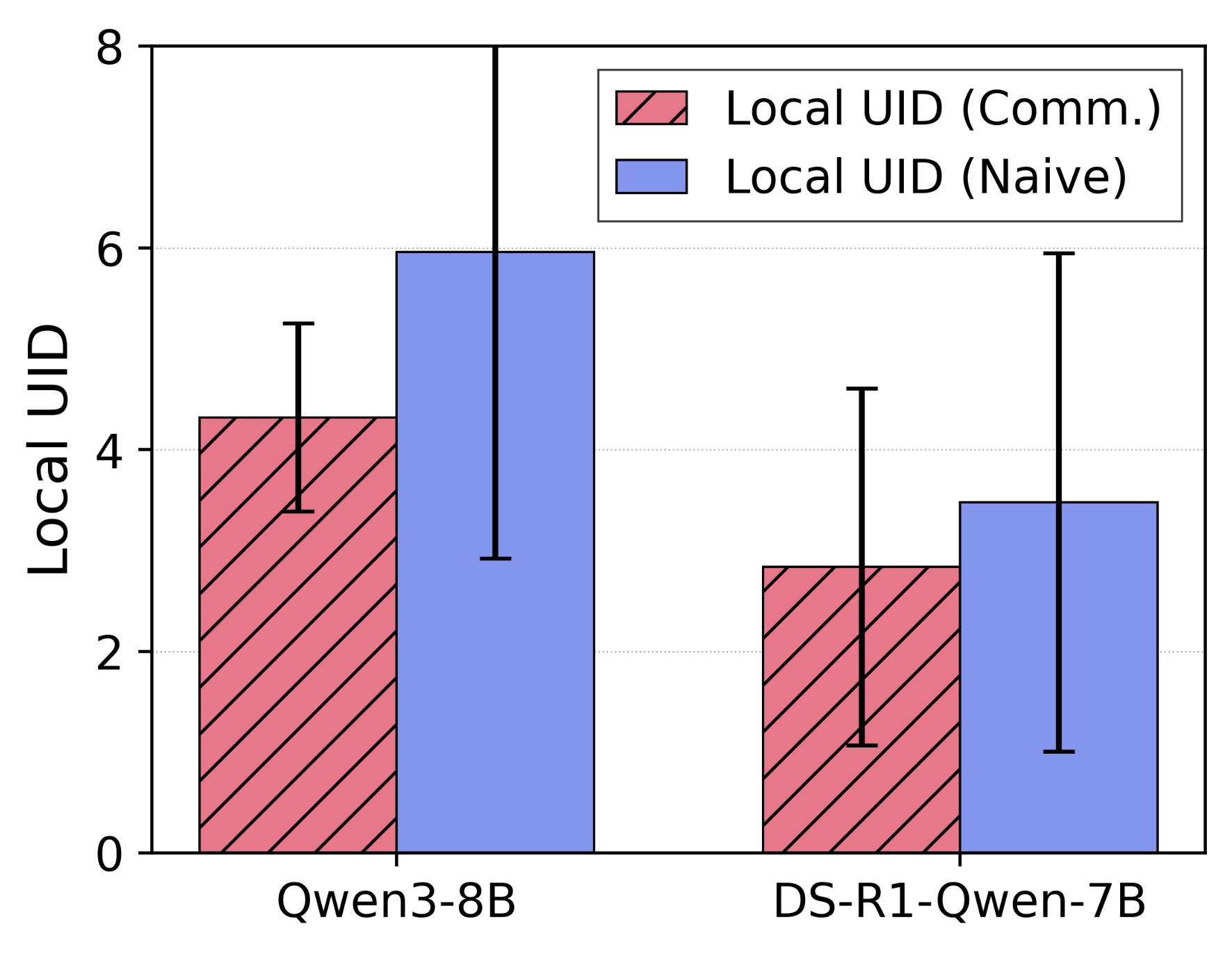}
        \caption{Local UID of comm vs.\ naive prompting}
        \label{fig:uid_local}
    \end{subfigure}

    \caption{
    Communication prompts induce higher global variance while yielding smoother local transitions,
    indicating more explanation-oriented pacing.
    }
    \vspace{-0.3in}
    \label{fig:uid_prompting}
\end{figure}

\section{Why Does UID Differ between Human Communication and LLM Reasoning?}

Our results show that high-quality LLM reasoning traces typically exhibit smooth local transitions in information density together with structured global non-uniformity. 
We conjecture that such deviation from the original UID hypothesis in psycholinguistics is due to different objectives between human communication and LLM reasoning rather than a deficiency in model behavior.

\begin{itemize}[leftmargin=3.0mm,topsep=3pt,itemsep=3pt,parsep=0pt]
    \item[$\circ$] \textbf{Human Communication: Normative, Listener-Optimized UID.} 
    In human language, UID is a \emph{normative} principle of communication. 
    Speakers assume the presence of a listener and aim to distribute information evenly over time to reduce processing difficulty under cognitive constraints. 
    Here, \textit{global uniformity is desirable}, as sharp information spikes can overload the listener. 
    \item[$\circ$] \textbf{LLM Reasoning: Instrumental, Computation-Driven UID.} 
    LLM reasoning traces, in contrast, are typically generated in a \emph{listener-free} setting. 
    CoT reasoning reflects an internal computational process rather than an act of communication. 
    Consequently, UID in LLMs is better understood as \emph{instrumental}: it characterizes how information is explored and evolved during problem solving, rather than how information is packaged for an external recipient. 
    Here, \textit{global non-uniformity} is not problematic but expected. 
    High-quality reasoning traces exhibit initial exploration with higher entropy, followed by consolidation and a final commitment phase where uncertainty collapses. 
    These transitions naturally produce uneven information distribution across steps, which is captured by higher global variance. 
    In contrast, incorrect traces tend to show noisy fluctuations without coherent phase structure.
\end{itemize} 
To verify the role of communicative objectives, we compare a conventional listener-free reasoning prompt with \textit{a new listener-aware (communication) prompt} (see Appendix~\ref{app:prompts}) that instruct the model to explain its reasoning to a listener. 
Under the same model, communication prompting consistently increases global UID, by approximately 6\% for Qwen3-8B and 11\% for DS-R1-Qwen-7B (Figure~\ref{fig:uid_prompting}, top), indicating more explanation-oriented pacing. 
Simultaneously, it reduces local UID, with about a 29\% decrease for Qwen3-8B and an 18\% decrease for DS-R1-Qwen-7B (Figure~\ref{fig:uid_prompting}, bottom), reflecting smoother step-to-step transitions with fewer abrupt spikes and falls.

\input{Tables/CommPrompting}

These findings clarify the role of UID as an evaluation signal. High-quality internal reasoning need not resemble human explanations, and penalizing global non-uniformity may harm reasoning assessment even if it improves readability. At the same time, local smoothness remains a shared indicator of coherence. Consistent with this interpretation, Table~\ref{tab:comm_vs_naive} shows that adopting a listener-aware, communicative style of reasoning leads to a performance decrease, highlighting that communicative optimality and internal reasoning effectiveness optimize for fundamentally different objectives. Overall, our suggested UID-based measures are best interpreted as internal, trace-level diagnostics of reasoning dynamics, rather than as proxies for human-likeness or communicative optimality.

\begin{AIbox}{Takeaway for Human vs. LLM Reasoning}
Human communication is listener-optimized and therefore favors global uniformity; LLM reasoning is computation-driven and naturally exhibits structured global non-uniformity from phase-based problem solving. Only local smoothness reliably indicates reasoning coherence.
\end{AIbox}


%% file: Tables/MainResults_ds.tex
\begin{table*}[t]
\caption{\textbf{Main results.} Performance on four math benchmarks (AIME2025, BRUMO2025, HMMT2025, and MinervaMath (MM)). Results are averaged over seeds 42, 1234, and 2025 with standard deviations. The best and second-best scores are highlighted in \textbf{bold} and \underline{underline}, respectively. Full results are in Appendix~\ref{app:main}.}
\centering

{\large
\resizebox{\textwidth}{!}{%
\begin{tabular}{@{}l|cccc|cccc|cccc@{}}
\toprule
 & \multicolumn{4}{c|}{\texttt{DS-R1-Distill-Qwen-7B}} 
 & \multicolumn{4}{c|}{\texttt{DS-R1-Distill-Llama-8B}}
 & \multicolumn{4}{c}{\texttt{Qwen3-8B}} \\
\cmidrule(lr){2-5} \cmidrule(lr){6-9} \cmidrule(lr){10-13}

Methods ($\downarrow$)
& AIME25 & BRUMO25 & HMMT25 & MM
& AIME25 & BRUMO25 & HMMT25 & MM
& AIME25 & BRUMO25 & HMMT25 & MM \\
\midrule

Mean Acc.
& \Large{0.40} & \Large{0.54} & \Large{0.24} & \Large{0.30}
& \Large{0.33} & \underline{\Large{0.40}} & \Large{0.20} & \Large{0.23}
& \Large{0.67} & \Large{0.68} & \Large{0.43} & \Large{\underline{0.34}} \\
\rowcolor{grayhighlight}
& \quad\quad{\footnotesize$\pm$0.02} & \quad\quad{\footnotesize$\pm$0.01} & \quad\quad{\footnotesize$\pm$0.00} & \quad\quad{\footnotesize$\pm$0.00}
& \quad\quad{\footnotesize$\pm$0.01} & \quad\quad{\footnotesize$\pm$0.02} & \quad\quad{\footnotesize$\pm$0.01} & \quad\quad{\footnotesize$\pm$0.00}
& \quad\quad{\footnotesize$\pm$0.01} & \quad\quad{\footnotesize$\pm$0.02} & \quad\quad{\footnotesize$\pm$0.01} & \quad\quad{\footnotesize$\pm$0.01} \\

Self-Cert.
& \Large{0.48} & \Large{0.52} & \underline{\Large{0.28}} & \Large{0.30}
& \Large{0.33} & \Large{0.34} & \Large{0.19} & \Large{0.22}
& \Large{0.63} & \Large{\textbf{0.71}} & \Large{\textbf{0.50}} & \Large{\underline{0.34}} \\
\rowcolor{grayhighlight}
& \quad\quad{\footnotesize$\pm$0.04} & \quad\quad{\footnotesize$\pm$0.04} & \quad\quad{\footnotesize$\pm$0.02} & \quad\quad{\footnotesize$\pm$0.00}
& \quad\quad{\footnotesize$\pm$0.00} & \quad\quad{\footnotesize$\pm$0.02} & \quad\quad{\footnotesize$\pm$0.02} & \quad\quad{\footnotesize$\pm$0.01}
& \quad\quad{\footnotesize$\pm$0.00} & \quad\quad{\footnotesize$\pm$0.02} & \quad\quad{\footnotesize$\pm$0.03} & \quad\quad{\footnotesize$\pm$0.01} \\

High Conf.
& \Large{0.48} & \Large{0.52} & \Large{0.27} & \Large{0.30}
& \Large{0.36} & \Large{0.36} & \Large{0.21} & \Large{\textbf{0.23}}
& \Large{0.60} & \Large{0.63} & \Large{0.42} & \Large{0.33} \\
\rowcolor{grayhighlight}
& \quad\quad{\footnotesize$\pm$0.04} & \quad\quad{\footnotesize$\pm$0.05} & \quad\quad{\footnotesize$\pm$0.00} & \quad\quad{\footnotesize$\pm$0.00}
& \quad\quad{\footnotesize$\pm$0.02} & \quad\quad{\footnotesize$\pm$0.03} & \quad\quad{\footnotesize$\pm$0.02} & \quad\quad{\footnotesize$\pm$0.01}
& \quad\quad{\footnotesize$\pm$0.00} & \quad\quad{\footnotesize$\pm$0.03} & \quad\quad{\footnotesize$\pm$0.06} & \quad\quad{\footnotesize$\pm$0.01} \\

Low Ent.
& \Large{0.48} & \underline{\Large{0.56}} & \Large{0.24} & \Large{0.30}
& \Large{0.34} & \Large{0.37} & \Large{0.19} & \Large{0.22}
& \Large{0.60} & \Large{0.64} & \Large{0.43} & \Large{0.33} \\
\rowcolor{grayhighlight}
& \quad\quad{\footnotesize$\pm$0.04} & \quad\quad{\footnotesize$\pm$0.05} & \quad\quad{\footnotesize$\pm$0.02} & \quad\quad{\footnotesize$\pm$0.00}
& \quad\quad{\footnotesize$\pm$0.04} & \quad\quad{\footnotesize$\pm$0.06} & \quad\quad{\footnotesize$\pm$0.02} & \quad\quad{\footnotesize$\pm$0.01}
& \quad\quad{\footnotesize$\pm$0.00} & \quad\quad{\footnotesize$\pm$0.04} & \quad\quad{\footnotesize$\pm$0.05} & \quad\quad{\footnotesize$\pm$0.01} \\

\midrule
\multicolumn{4}{l}{UID Metrics ($\downarrow$)} \\
\midrule

Loc. non-uni
& \Large{0.27} & \Large{0.39} & \Large{0.18} & \Large{0.26}
& \Large{0.36} & \Large{0.36} & \Large{0.20} & \textbf{\Large{0.23}}
& \Large{0.63} & \Large{0.63} & \Large{0.40} & \Large{0.34} \\
\rowcolor{grayhighlight}
& \quad\quad{\footnotesize$\pm$0.00} & \quad\quad{\footnotesize$\pm$0.07} & \quad\quad{\footnotesize$\pm$0.04} & \quad\quad{\footnotesize$\pm$0.01} 
& \quad\quad{\footnotesize$\pm$0.04} & \quad\quad{\footnotesize$\pm$0.03} & \quad\quad{\footnotesize$\pm$0.03} & \quad\quad{\footnotesize$\pm$0.01}
& \quad\quad{\footnotesize$\pm$0.00} & \quad\quad{\footnotesize$\pm$0.07} & \quad\quad{\footnotesize$\pm$0.03} & \quad\quad{\footnotesize$\pm$0.01} \\

\rowcolor{lightblue}
Loc. uni
& \Large{\textbf{0.53}} & \Large{0.56} & \Large{\textbf{0.30}} & \Large{\textbf{0.31}}
& \Large{\textbf{0.39}} & \Large{\textbf{0.44}} & \Large{\underline{0.24}} & \Large{\textbf{0.23}}
& \Large{\underline{0.69}} & \Large{\underline{0.70}} & \Large{\underline{0.48}} & \Large{\underline{0.34}} \\
\rowcolor{lightblue}

& \quad\quad{\footnotesize$\pm$0.06} & \quad\quad{\footnotesize$\pm$0.02} & \quad\quad{\footnotesize$\pm$0.00} & \quad\quad{\footnotesize$\pm$0.01}

& \quad\quad{\footnotesize$\pm$0.04} & \quad\quad{\footnotesize$\pm$0.06} & \quad\quad{\footnotesize$\pm$0.04} & \quad\quad{\footnotesize$\pm$0.01}

& \quad\quad{\footnotesize$\pm$0.03} & \quad\quad{\footnotesize$\pm$0.03} & \quad\quad{\footnotesize$\pm$0.04} & \quad\quad{\footnotesize$\pm$0.01} \\

\rowcolor{lightblue}
Glob. non-uni
& \Large{\underline{0.52}} & \Large{\textbf{0.64}} & \Large{0.26} & \Large{\underline{0.30}}
& \Large{\underline{0.37}} & \Large{0.36} & \Large{0.18} & \Large{0.21}
& \Large{\textbf{0.70}} & \Large{0.61} & \Large{0.47} & \Large{0.33} \\
\rowcolor{lightblue}
& \quad\quad{\footnotesize$\pm$0.08} & \quad\quad{\footnotesize$\pm$0.05} & \quad\quad{\footnotesize$\pm$0.04} & \quad\quad{\footnotesize$\pm$0.01}

& \quad\quad{\footnotesize$\pm$0.03} & \quad\quad{\footnotesize$\pm$0.04} & \quad\quad{\footnotesize$\pm$0.04} & \quad\quad{\footnotesize$\pm$0.00}

& \quad\quad{\footnotesize$\pm$0.00} & \quad\quad{\footnotesize$\pm$0.06} & \quad\quad{\footnotesize$\pm$0.03} & \quad\quad{\footnotesize$\pm$0.01} \\

Glob. uni
& \Large{0.33} & \Large{0.44} & \Large{0.16} & \Large{0.28}
& \Large{0.27} & \underline{\Large{0.40}} & \Large{\textbf{0.27}} & \Large{0.23}
& \Large{0.66} & \Large{0.68} & \Large{0.41} & \Large{\textbf{0.34}} \\
\rowcolor{grayhighlight}
& \quad\quad{\footnotesize$\pm$0.00} & \quad\quad{\footnotesize$\pm$0.02} & \quad\quad{\footnotesize$\pm$0.04} & \quad\quad{\footnotesize$\pm$0.00}
& \quad\quad{\footnotesize$\pm$0.12} & \quad\quad{\footnotesize$\pm$0.06} & \quad\quad{\footnotesize$\pm$0.03} & \quad\quad{\footnotesize$\pm$0.01}
& \quad\quad{\footnotesize$\pm$0.02} & \quad\quad{\footnotesize$\pm$0.07} & \quad\quad{\footnotesize$\pm$0.02} & \quad\quad{\footnotesize$\pm$0.02} \\

\bottomrule
\end{tabular}}

\vspace{-0.1in}
}
\label{tab:mainresults_ds}
\end{table*}

%% file: Tables/ScaleModel.tex
\begin{table}[t]
\caption{\textbf{Model size analysis.} Performance of \texttt{Qwen3} models (1.7B, 4B, 8B) on AIME2025. Results are averaged over seeds 42, 1234, and 2025. The best and second-best scores are highlighted in \textbf{bold} and \underline{underline}, respectively. Full results are in Appendix~\ref{app:model-size}.}
\vspace{-0.1in}
\begin{center}
\footnotesize
\resizebox{0.9\linewidth}{!}{
\begin{tabular}{l|c|c|c}
\toprule
{Methods ($\downarrow$)} & \texttt{Qwen3-1.7B} & \texttt{Qwen3-4B} & \texttt{Qwen3-8B} \\ \midrule

Mean Acc. 
& 0.35 & 0.65 & 0.67 \\

Self-Cert.
& \textbf{0.45} & \textbf{0.73} & 0.63 \\

High Conf. 
& 0.37 & 0.62 & 0.60 \\

Low Ent. 
& 0.37 & 0.63 & 0.60 \\

\midrule
\multicolumn{4}{l}{UID Metrics ($\downarrow$)} \\
\midrule

Loc. non-uni
& 0.24 & 0.54 & 0.63 \\

\rowcolor{lightblue}
Loc. uni
& \underline{0.41} & \underline{0.69} & \underline{0.69} \\

\rowcolor{lightblue}
Glob. non-uni
& 0.37 & 0.66 & \textbf{0.70} \\

Glob. uni
& 0.33 & 0.67 & 0.66 \\

\bottomrule
\end{tabular}
}
\end{center}
\vspace{-0.2in}
\label{tab:scalemodel}
\end{table}

%% file: Tables/NonMath.tex
\begin{table}[t]
    \caption{\textbf{Reasoning performance outside the math domain.}
    The best and second-best scores are highlighted in \textbf{bold} and \underline{underline}, respectively.}
    \label{tab:nomath}
    \centering
    
    \begin{adjustbox}{width=\columnwidth}
    \begin{tabular}{l|ccc|ccc}
        \toprule
        & \multicolumn{3}{c}{\texttt{DS-R1-Distill-Llama-8B}} 
        & \multicolumn{3}{c}{\texttt{DS-R1-Distill-Qwen-7B}} \\
        \cmidrule(lr){2-4} \cmidrule(lr){5-7}
        Methods ($\downarrow$) & GPQA-D & LSAT-AR & LSAT-LR & GPQA-D & LSAT-AR & LSAT-LR \\
        \midrule
        Mean Acc.      & 0.48 & 0.43 & 0.52 & 0.48 & 0.55 & 0.50 \\
        Self-Cert.     & 0.51 & 0.44 & 0.52 & \textbf{0.51} & 0.51 & 0.49 \\
        High Conf.     & \textbf{0.53} & 0.43 & 0.51 & 0.46 & 0.51 & 0.51 \\
        Low Ent.       & 0.49 & 0.42 & 0.52 & 0.48 & 0.51 & \underline{0.52} \\
        \midrule
        \multicolumn{7}{l}{{UID Metrics} ($\downarrow$)} \\
        \midrule
        Loc. non-uni   & 0.44 & 0.39 & 0.49 & 0.43 & 0.48 & 0.46 \\
        \rowcolor{lightblue}
        Loc. uni       & \underline{0.52} & \underline{0.46} & \underline{0.53} & \textbf{0.51} & \textbf{0.62} & 0.49 \\
        \rowcolor{lightblue}
        Glob. non-uni  & 0.47 & \textbf{0.49} & \textbf{0.54} & 0.50 & \textbf{0.62} & \textbf{0.53} \\
        Glob. uni      & 0.44 & 0.40 & 0.50 & 0.46 & 0.46 & 0.46 \\
        \bottomrule
    \end{tabular}
    \end{adjustbox}
    \vspace{-0.1in} 
\end{table}

%% file: Tables/ScaleSample.tex
\begin{table}[t]
\caption{\textbf{Sample size analysis}. Sample by 3, 5, 10 on AIME using \texttt{Qwen3-8B}. Results are averaged over seeds 42, 1234, and 2025 with standard deviations. The best and second-best scores are highlighted in \textbf{bold} and \underline{underline}, respectively. Full results are in Appendix~\ref{app:sample-scale}.}
\begin{center}
\resizebox{1.0\linewidth}{!}{
\begin{tabular}{l|ccc}
\toprule
{Methods ($\downarrow$)} & Sample by 3 & Sample by 5 & Sample by 10 \\
\midrule
Mean Acc.     
& 0.67
& 0.67
& 0.68 \\

Self-Cert.  
& {0.70}
& 0.63
& 0.62 \\

High Conf.       
& {0.63}
& 0.60
& 0.57 \\

Low Ent.         
& {0.63}
& 0.60
& 0.56 \\
\midrule
UID Metrics ($\downarrow$) \\
\midrule
Loc. non-uni
& 0.63
& 0.63
& 0.53 \\

\rowcolor{lightblue}
Loc. uni
& \textbf{0.73}
& \underline{0.69}
& \underline{0.72} \\

\rowcolor{lightblue}
Glob. non-uni     
& \underline{0.70}
& \textbf{0.70}
& \textbf{0.70} \\

Glob. uni      
& {0.70}
& 0.66
& 0.63 \\

\bottomrule
\end{tabular}
}
\end{center}
\vspace{-0.2in}
\label{tab:scalesample}
\end{table}

%% file: Tables/CommPrompting.tex
\begin{table}[t]
\centering
\small
\begin{tabular}{lcc}
\toprule
\textbf{Model} & \textbf{Comm.} & \textbf{Naive} \\
\midrule
\texttt{DS-R1-Distill-Qwen-7B} & 0.37 & \textbf{0.41} \\
\texttt{Qwen3-8B}                & 0.63 & \textbf{0.67} \\
\bottomrule
\end{tabular}
\caption{Performance under comm. and naive settings.}
\vspace{-0.2in}
\label{tab:comm_vs_naive}
\end{table}

%% file: 2_related_work.tex
\section{Related Work}

\paragraph{Fragility of CoT and the role of individual reasoning steps.}CoT prompting improves reasoning but remains fragile ~\citep{wei2023chainofthoughtpromptingelicitsreasoning, zhao2025chainofthoughtreasoningllmsmirage, kojima2023largelanguagemodelszeroshot, chae2023dialoguechainofthoughtdistillationcommonsenseaware, chen2023programthoughtspromptingdisentangling}. 
Small, seemingly irrelevant perturbations in the reasoning chain can sharply reduce accuracy \citep{mirzadeh2025gsmsymbolicunderstandinglimitationsmathematical,tang2023largelanguagemodelsincontext}, suggesting that models often produce the appearance of reasoning rather than logically sound traces \citep{shojaee2025illusionthinkingunderstandingstrengths}. 
Moreover, longer reasoning steps do not necessarily reflect the true difficulty of the problem, and many intermediate steps can be altered or even removed without changing the final answer \citep{lanham2023measuringfaithfulnesschainofthoughtreasoning}. 
This raises doubts about the necessity and faithfulness of these step-by-step explanations.
Another line of research takes a different perspective: rather than viewing all steps as equally important, it suggests that a small subset of pivotal steps within CoT traces disproportionately drives predictions \citep{bogdan2025thoughtanchorsllmreasoning}.
Attribution methods and their frameworks identify and highlight these critical steps, emphasizing the need to understand how individual steps shape outcomes \citep{golovneva2023roscoesuitemetricsscoring, wu2023analyzingchainofthoughtpromptinglarge, bigelow2024forkingpathsneuraltext}. 
Despite these advances, prior works have no clear interpretations of what constitutes as a truly good reasoning pattern.

\paragraph{Intrinsic signals in LLM reasoning.}Research on LLM reasoning has increasingly turned to internal model signals to gain insight into how reasoning unfolds \citep{zhao2025learningreasonexternalrewards, zhang2025rightquestionhalfanswer}.
Many approaches use these signals to improve performance \citep{zuo2025ttrltesttimereinforcementlearning}, such as self-certainty \citep{kang2025scalablebestofnselectionlarge, zhao2025learningreasonexternalrewards}, or confidence \citep{jang2025selftraininglargelanguagemodels} to refine outputs, or using entropy-based measures to encourage diverse reasoning paths \citep{zhang2025rightquestionhalfanswer, agarwal2025unreasonableeffectivenessentropyminimization, gao2025oneshotentropyminimization, lee2025trainingfreellmverificationrecycling, li2025compressingchainofthoughtllmsstep, zhou-etal-2023-inform}. However, these methods largely treat internal signals as heuristics for guiding or controlling reasoning, without providing a principled account of why certain reasoning traces are more coherent than others. In contrast, we ground our analysis in the long-standing psycholinguistic theory of Uniform Information Density (UID) hypothesis, which offers a theoretical lens for understanding how information is introduced, transformed, and propagated through reasoning. Our perspective on information flow in reasoning traces moves beyond performance heuristics, emphasizing structural properties that characterize coherence and clarify how reasoning differs from human communication.



%% file: 6_conclusion.tex
\section{Conclusion}
This paper revisits the long-standing Uniform Information Density (UID) hypothesis in the context of large language model reasoning. 
By shifting the focus from output-level correctness to step-level information flow, we demonstrate that entropy-uniformity serves as a meaningful indicator of reasoning quality. 
Our analysis reveals that coherent reasoning traces tend to distribute information more evenly across steps, while disfluent traces exhibit sharp entropy fluctuations. 
On the other hand, non-uniform traces at the global level with higher variance leads to a more coherent reasoning trace. 
These findings bridge psycholinguistic theory with computational analysis, providing a new lens for interpreting model reasoning beyond performance metrics. 
Ultimately, our work suggests that UID-inspired measures can guide the design of more interpretable and trustworthy reasoning systems.

\section*{Limitations}
While our study highlights the importance of uniformity in identifying coherent reasoning traces, several limitations and open questions remain.

First, our analysis is primarily restricted to structured reasoning datasets, which may not fully capture model behavior in broader settings such as open-ended dialogue or interactive communication. Notably, our additional prompting analysis reveals a systematic difference between listener-free internal reasoning and listener-oriented explanation: in naive settings, models tend to produce abrupt transitions and uneven intermediate steps with relatively low global variation, whereas introducing a notional listener shifts generation toward smoother local transitions and higher global variance. These observations support our interpretation that the divergence between human and model UID arises from a structural difference between internal reasoning and communicative reasoning. While this analysis provides targeted empirical evidence for a contrast that, to our knowledge, has not been explored previously, future work is needed to evaluate whether this phenomenon generalizes across domains, modalities, and interaction settings.

Second, our methodology focuses on token- and step-level entropy dynamics as proxies for information allocation during generation. 
Although this perspective offers a tractable and informative view of reasoning structure, it does not provide a mechanistic explanation of why such differences emerge. In particular, we do not explicitly connect the observed UID patterns to the underlying properties of autoregressive decoding, internal information allocation, or training objectives of large language models. A deeper mechanistic account linking generation dynamics to these behavioral signatures would further enrich this line of work and remains an important direction for future research.

\section*{Broader Impact and Ethical Considerations}
This work contributes to improving the reliability and interpretability of LLM reasoning by introducing information-theoretic diagnostics that characterize how information is distributed and evolves within individual reasoning traces. By treating reasoning as an internal information-flow process rather than solely on final answers or cross-sample agreement, UID-based metrics provide a complementary, sample-efficient signal for identifying coherent versus unstable reasoning trajectories, which may support safer and more trustworthy deploymnet of LLMs in settings such as education, scientific problem-solving, and decision support. At the same time, these metrics are not guarantees of correctness and should not be interpreted as normative standards of good or human-like reasoning. Misuse may arise if UID scores are treated as hard filters, certification signals, or stylistic constraints, potentially leading to overconfidence in incorrect outputs or incentivizing superficial optimization of reasoning traces without genuine improvements in reasoning ability. Moreover, as our analysis shows that effective LLM reasoning diverges from human communicative UID patterns, applying these metrics to evaluate human-facing explanations or across domains beyond those studied may be inappropropriate. We therefore emphasize the UID-based measures are best understood as diagnostic tools operating within a broader evaluation framework that includes answer verification, aggregation methods, and human oversight. Responsible use requires clear communication of their scope, limitations, and probabilistic nature, as well as further study of their behavior across domains, languages, and interaction settings.

\section*{Acknowledgments}
We thank Ilgee Hong at the Georgia Institute of Technology for helpful discussions, and Chanjoo Jung at Yonsei University for valuable feedback on writing. We also thank Oneline AI for providing GPU resources that supported this research. Minju and Jaehyung are affiliated with the Department of Artificial Intelligence at Yonsei University. This research was supported in part by Institute for Information \& communications Technology Planning \& Evaluation (IITP) grant funded by the Korea government (MSIT) (No. RS-2020-II201361, Artificial Intelligence Graduate School Program (Yonsei University); No. RS-2025-25442405, Development of a Self-Learning World Model-Based AGI System for Hyperspectral Imaging.

%% file: 7_appendix.tex
\newpage
\section{Justification of Segmentation Strategy and Robustness Analyses}
\label{app:segmentation_robustness}

\subsection{Justification of newline-based segmentation}

We segment reasoning traces using paragraph boundaries defined by \texttt{\textbackslash n\textbackslash n}. This heuristic aligns with how large language models structure free-form reasoning and is widely adopted in prior work. While explicit empirical validation of step segmentation strategies remains limited, Lightman et~al.~\citep{lightman2023letsverifystepstep} employ newline-separated steps in the PRM800K dataset, where large-scale human annotation suggests that paragraph boundaries often align with intuitive subgoal transitions. 

Several recent studies on step-wise reasoning and process-level supervision similarly adopt newline- or paragraph-based segmentation schemes~\citep{feng2025stepbystepreasoningmathproblems,yang2025errorprocessrewardmodels}, indicating that this approach is a practical and commonly accepted default. Importantly, our method does not rely on the optimality of any single segmentation strategy; instead, we verify robustness to alternative stepwise segmentation choices in the following sections.

\input{Tables/APP_table6}
\input{Tables/APP_table7}

\subsection{Robustness to Fixed Token Window Segmentation}

To evaluate robustness to non-semantic step definitions, we apply a coarse fixed-length token window segmentation (2048 tokens per step) following ~\citet{fu2025deepthinkconfidence}. Table~\ref{tab:fixed_window} reports UID-conditioned performance across AIME, HMMT, and BRUMO for Qwen3-8B, DeepSeek-R1-Distill-Qwen-7B, and DeepSeek-R1-Distill-Llama-8B (seed 42).

Across nearly all model–dataset pairs, the expected ordering patterns persist: 
\textit{High UID (3$\sigma$) $\leq$ Low UID (3$\sigma$)} and 
\textit{High UID (var) $\geq$ Low UID (var)}. 
While we observe an isolated deviation for Qwen3-8B on BRUMO under variance-based grouping, the overall trend remains consistent, with local and global UID signals continuing to distinguish higher- and lower-quality reasoning traces.

\subsection{Robustness to Semantic Segmentation}

We further evaluate robustness under semantic segmentation, where steps are defined using discourse-level markers such as \emph{But}, \emph{Wait}, and \emph{Alternatively}, following the classifications mentioned in ~\citet{aggarwal2025l1controllinglongreasoning}. As shown in Table~\ref{tab:semantic_seg}, most settings preserve the expected UID-based orderings under both 3$\sigma$- and variance-based groupings.

We observe minor localized deviations, such as Qwen3-8B on AIME under the 3$\sigma$ grouping and DeepSeek-R1-Distill-Qwen-7B on HMMT under variance-based grouping. However, these deviations are not systematic and do not alter the broader trend: across AIME, HMMT, and BRUMO, and across all three model families, High UID (3$\sigma$) rarely outperforms Low UID (3$\sigma$), while High UID (var) generally exceeds Low UID (var). 

These results indicate that although semantic segmentation introduces slightly more variability—likely due to sensitivity in defining discourse boundaries—the core UID trends remain stable across segmentation strategies.

\section{What Entropy Does and Does Not Measure} In this work, entropy is used as a tractable proxy for step-level information density, not as a direct measure of semantic progress, factual correctness, or logical soundness. High entropy indicates that the model is distributing probability mass across multiple plausible next-token continuations, while low entropy indicates a more concentrated predictive distribution. However, either regime may arise in both successful and unsuccessful reasoning: elevated entropy can reflect productive exploration of alternatives, but it can also reflect confusion or hallucination; conversely, low entropy can indicate confident convergence, but also premature commitment to an incorrect path. For this reason, we do not interpret entropy magnitude in isolation. Instead, our hypothesis is that the structure of entropy across steps—whether uncertainty evolves smoothly and resolves in an organized way—provides a more reliable signal of reasoning quality than raw entropy alone.

\section{Details of Preliminary Analyses}
\subsection{Empirical analyses on various internal signals as a proxy for $ID_i$}\label{app:empirical}
We consider three metrics as a proxy for information density $ID_i$: (1) log-probability $LP_i$ as a confidence signal, composed from the average token log-probability over step $i$, (2) entropy $H_i$ as an uncertainty signal, and (3) confidence gap $D_i$ as divergence signal defined as the difference between the log-probability of the current and the previous steps.

Formally, log-probability $LP_i$ of a step is the average token log-probability over step $i$:
\[
LP_i = \frac{1}{b_i - a_i + 1} \sum_{t=a_i}^{b_i} \ell_t.
\]

While token-level entropy $H_t$ as
\[
H_t = - \sum_{v \in V} p_t(v) \log p_t(v),
\]

step-level entropy $H_i$ is defined as
\[
H_i = \frac{1}{b_i - a_i + 1} \sum_{t=a_i}^{b_i} H_t
\]

Log-probability gap $D_i$ is defined as
\[
D_i = LP_i - LP_{i-1}
\]

\begin{figure}[htbp!]
    \centering
    \begin{subfigure}[t]{\linewidth}
        \centering
        \includegraphics[width=\linewidth]{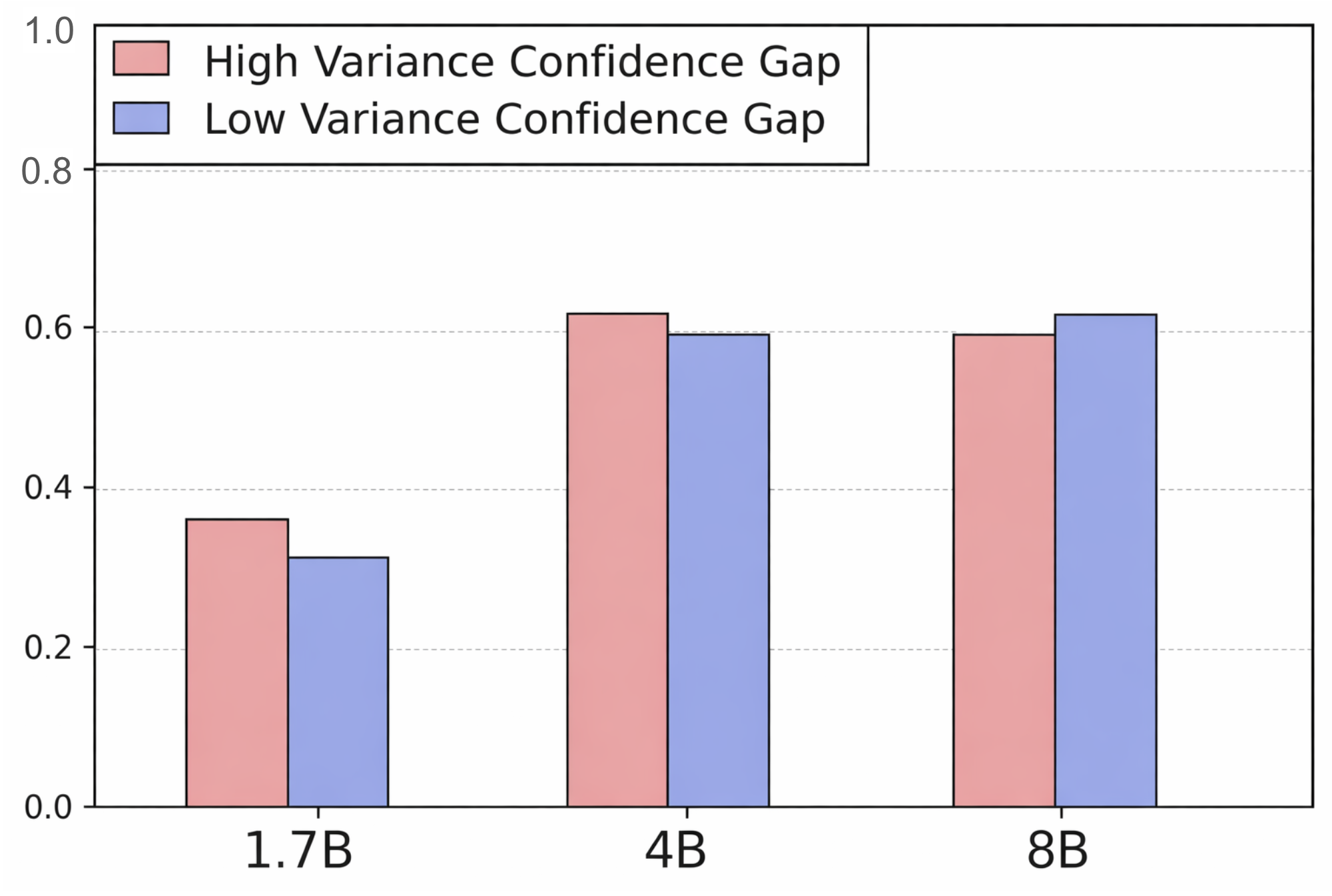}
        \caption{Acc. of traces with max vs. min var in confidence gap}
        \label{fig:confidence_gap}
    \end{subfigure}
    
    \begin{subfigure}[t]{\linewidth}
        \centering
        \includegraphics[width=\linewidth]{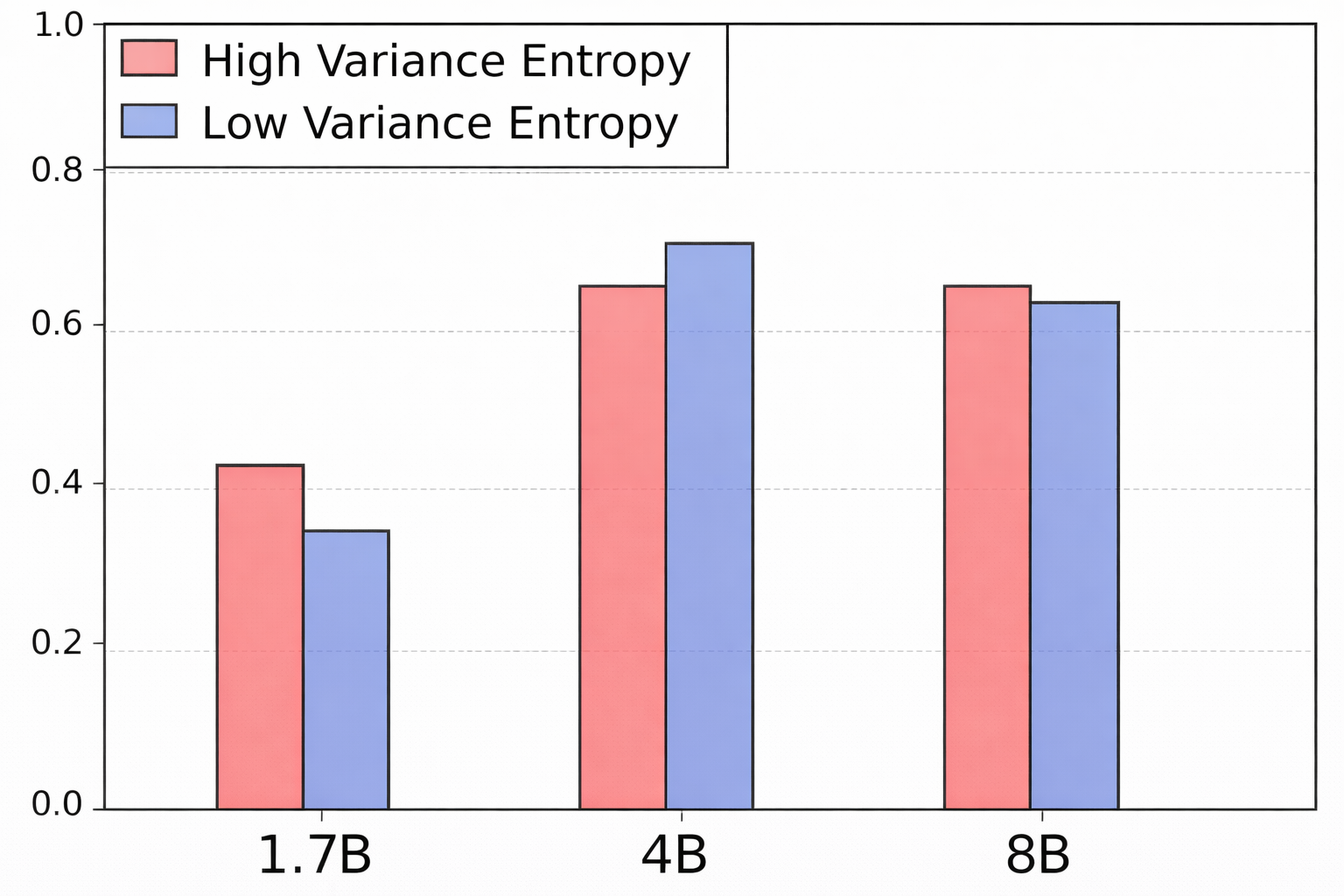}
        \caption{Acc. of traces with max vs. min var in entropy}
        \label{fig:entropy}
    \end{subfigure}

    \begin{subfigure}[t]{\linewidth}
        \centering
        \includegraphics[width=\linewidth]{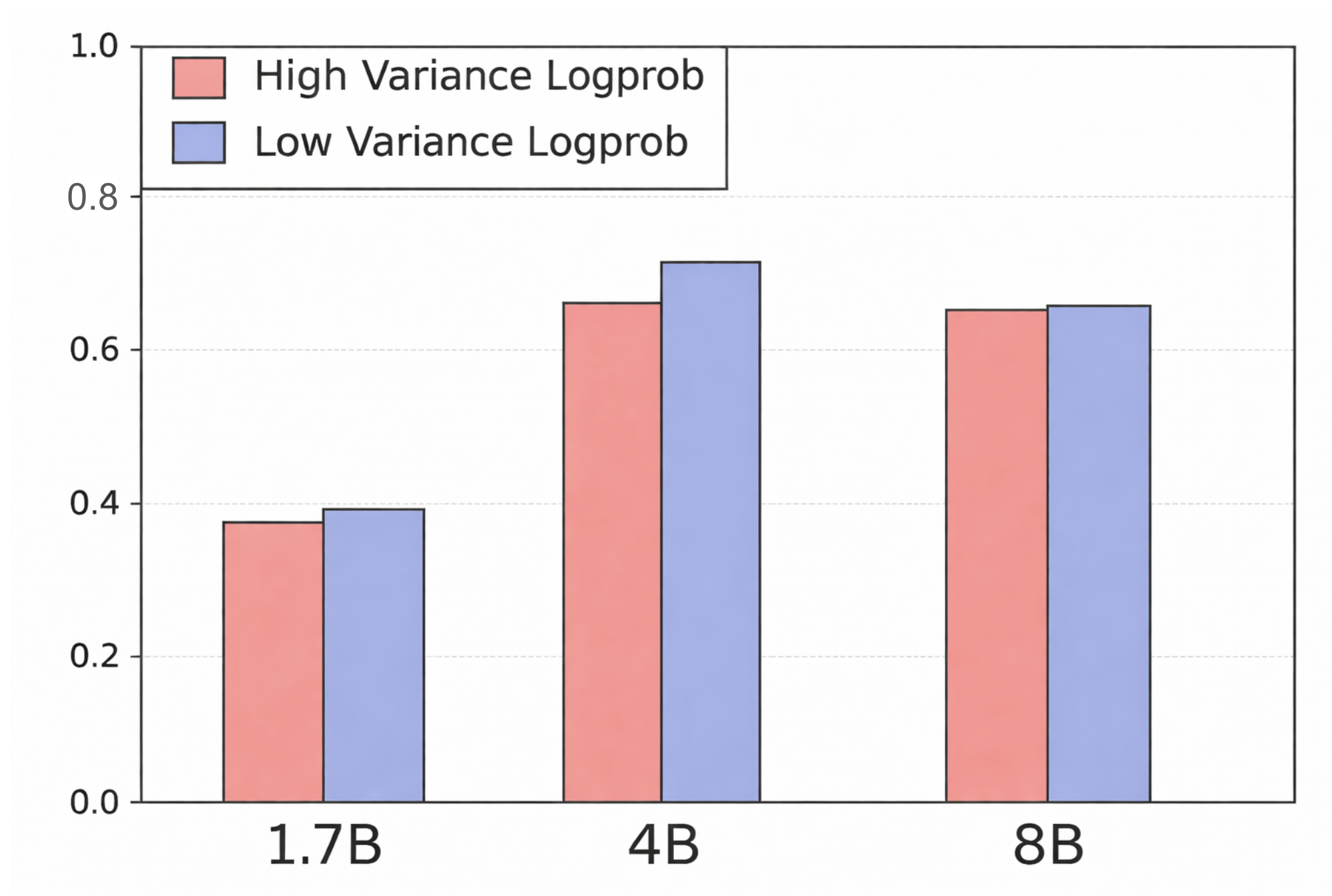}
        \caption{Acc. of traces with max vs. min var in log-probability}
        \label{fig:logprob}
    \end{subfigure}

    \caption{Empirical results on AIME2025 show that entropy-uniformity is most effective for identifying sound reasoning traces.}
    \label{fig:uid-selection}
\end{figure}

We calculate variance over reasoning traces where each $LP_i$, $H_i$, and $D_i$ is used as a proxy of $ID_i$, and calculate the accuracy of the traces with the highest and lowest variance (\ie, the degree of global uniformity). Results in Figure~\ref{fig:uid-selection} reveal that entropy measures the difference between highest and lowest trends more profoundly than others across various models. 

\subsection{Trace-level Reasoning Step Analyses by Answer Correctness}\label{app:trace}
\begin{figure*}[t]
\centering
\includegraphics[ width=\textwidth]{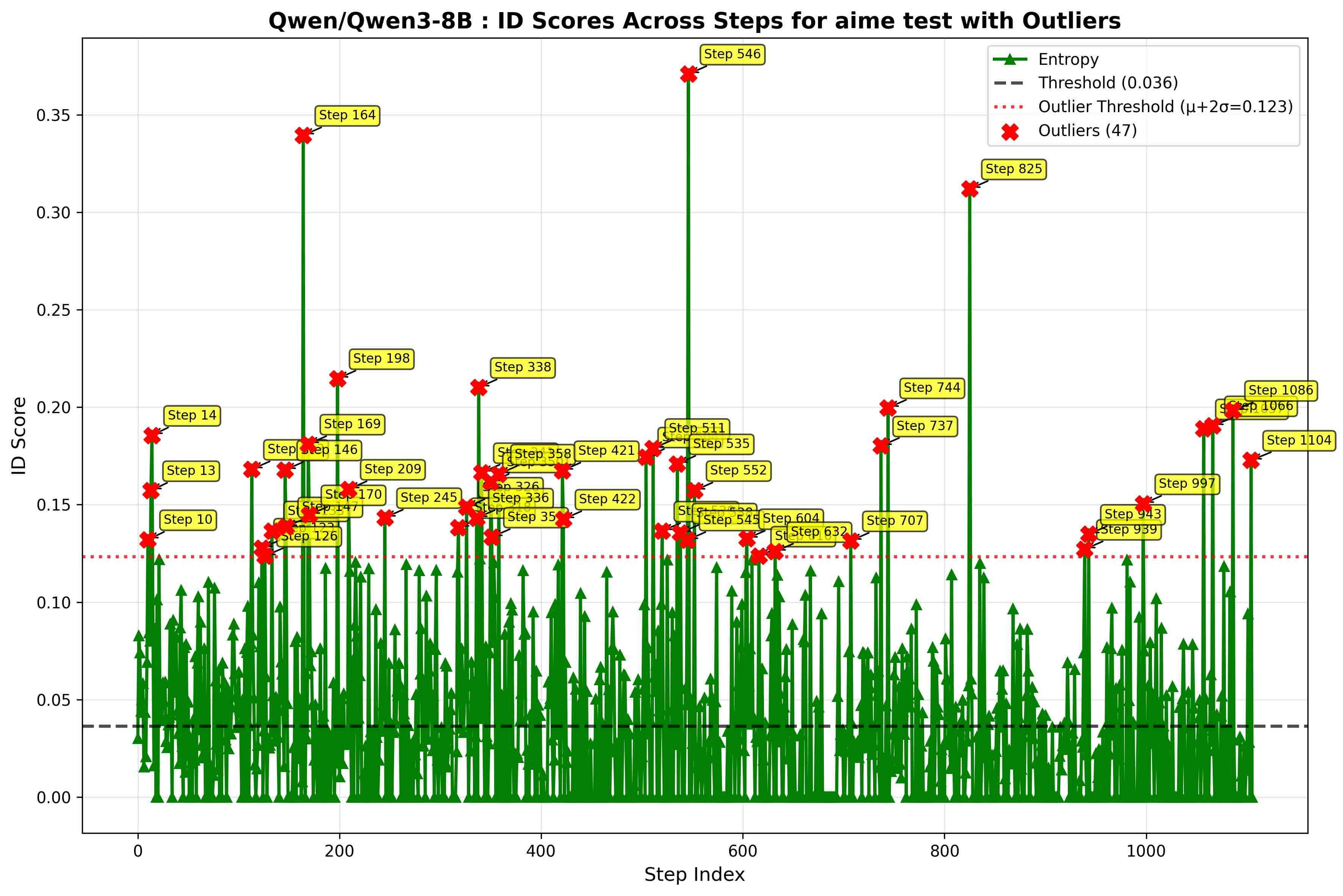}
\vspace{-0.1in}
\caption{Q6 — Correct Trace Visualization}
\vspace{-0.1in}
\label{fig:q6_correct_vis}
\end{figure*}

\begin{figure*}[t]
\centering
\includegraphics[width=\textwidth]{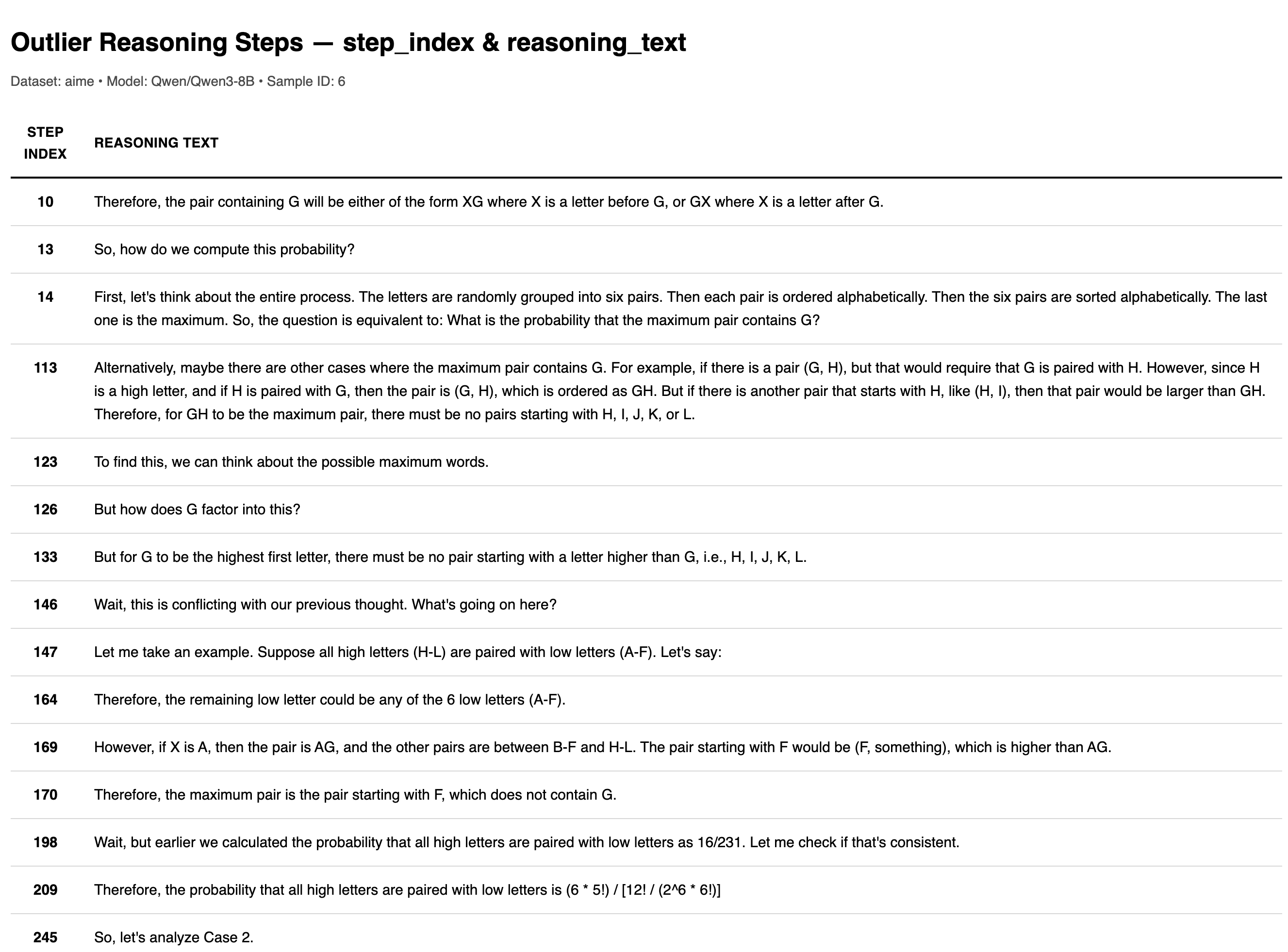}
\vspace{-0.1in}
\caption{Q6 — Correct Trace Text}
\vspace{-0.1in}
\label{fig:q6_correct_text}
\end{figure*}

\begin{figure*}[t]
\centering
\includegraphics[width=\textwidth]{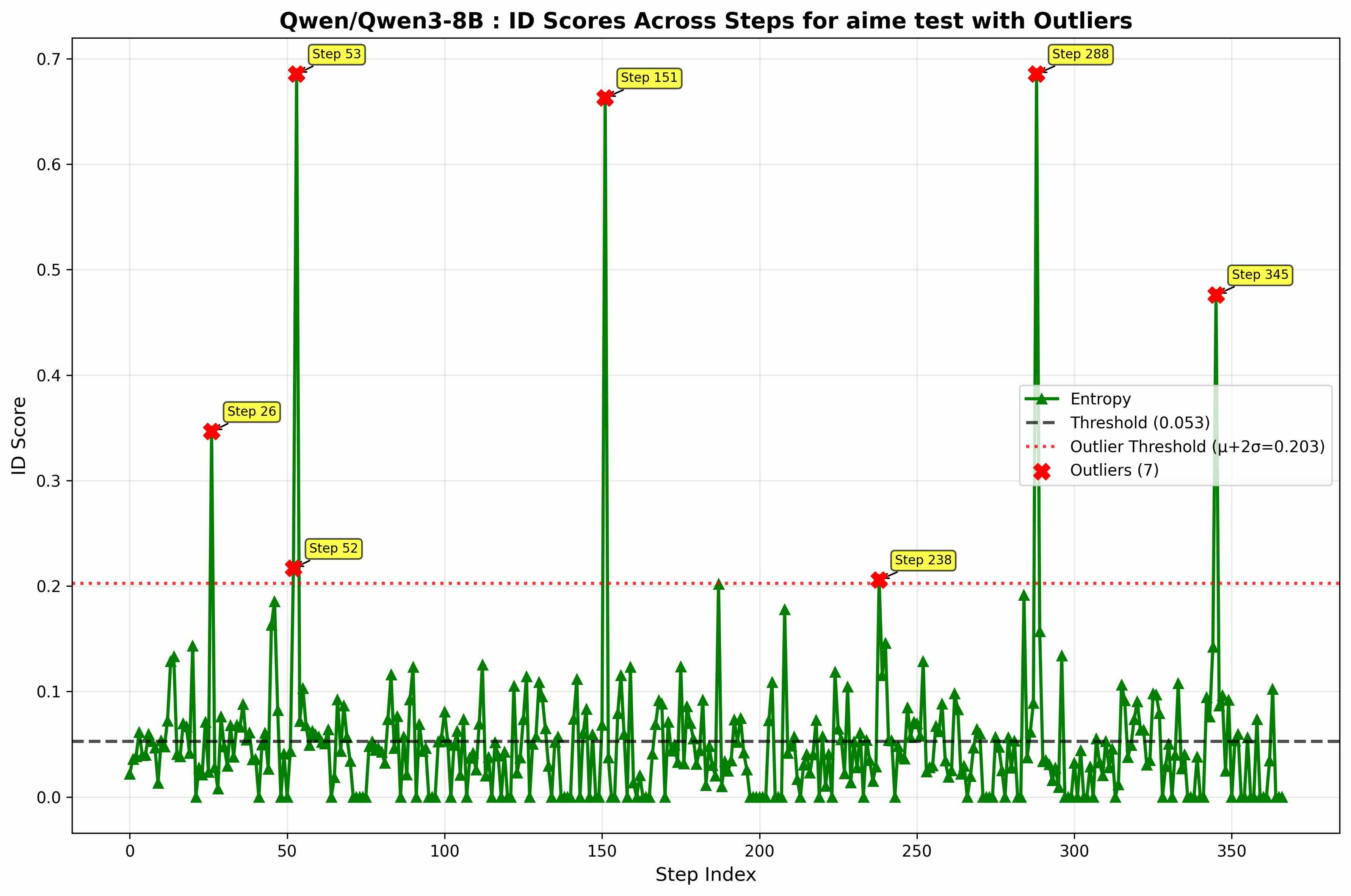}
\vspace{-0.1in}
\caption{Q6 — Incorrect Trace Visualization}
\vspace{-0.1in}
\label{fig:q6_incorrect_vis}
\end{figure*}

\begin{figure*}[t]
\centering
\includegraphics[width=\textwidth]{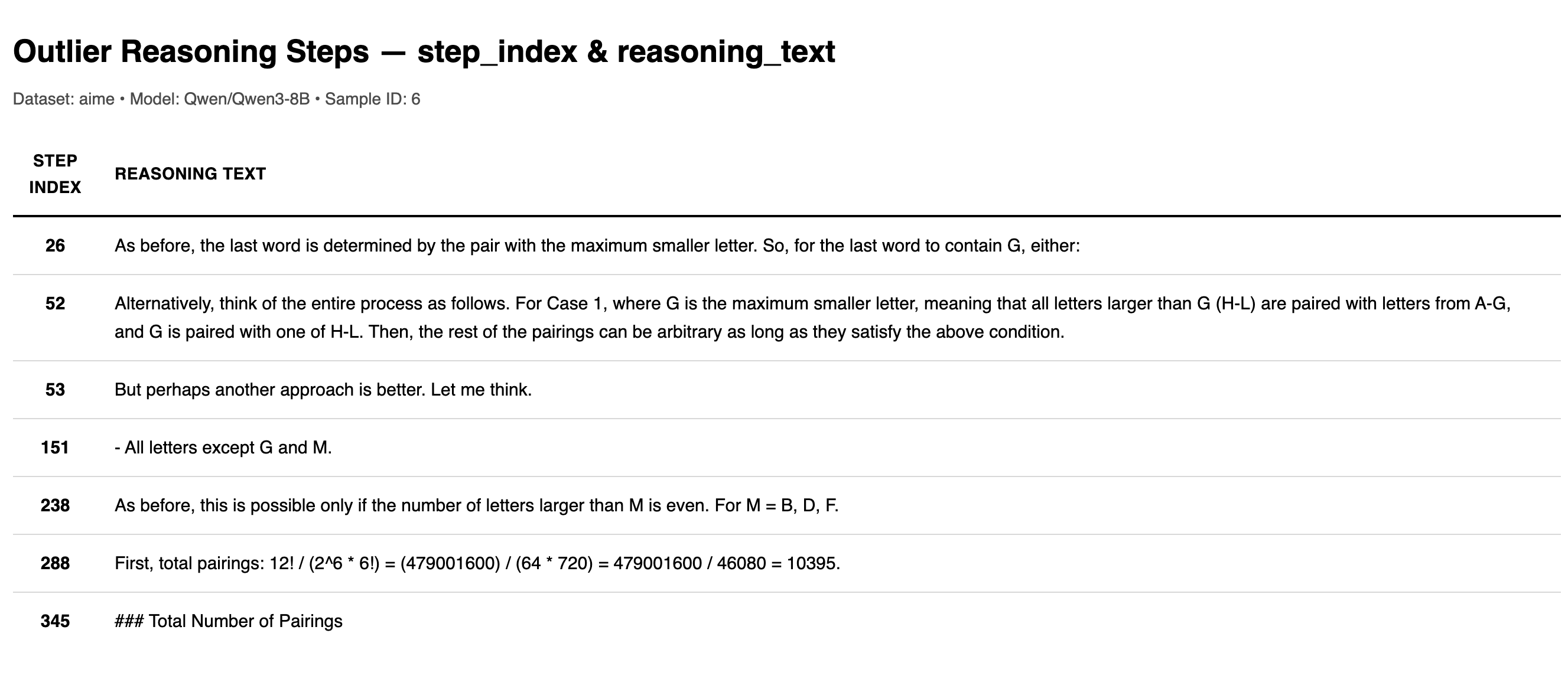}
\vspace{-0.1in}
\caption{Q6 — Incorrect Trace Text}
\vspace{-0.1in}
\label{fig:q6_incorrect_text}
\end{figure*}

\begin{figure*}[t]
\centering
\includegraphics[width=\textwidth]{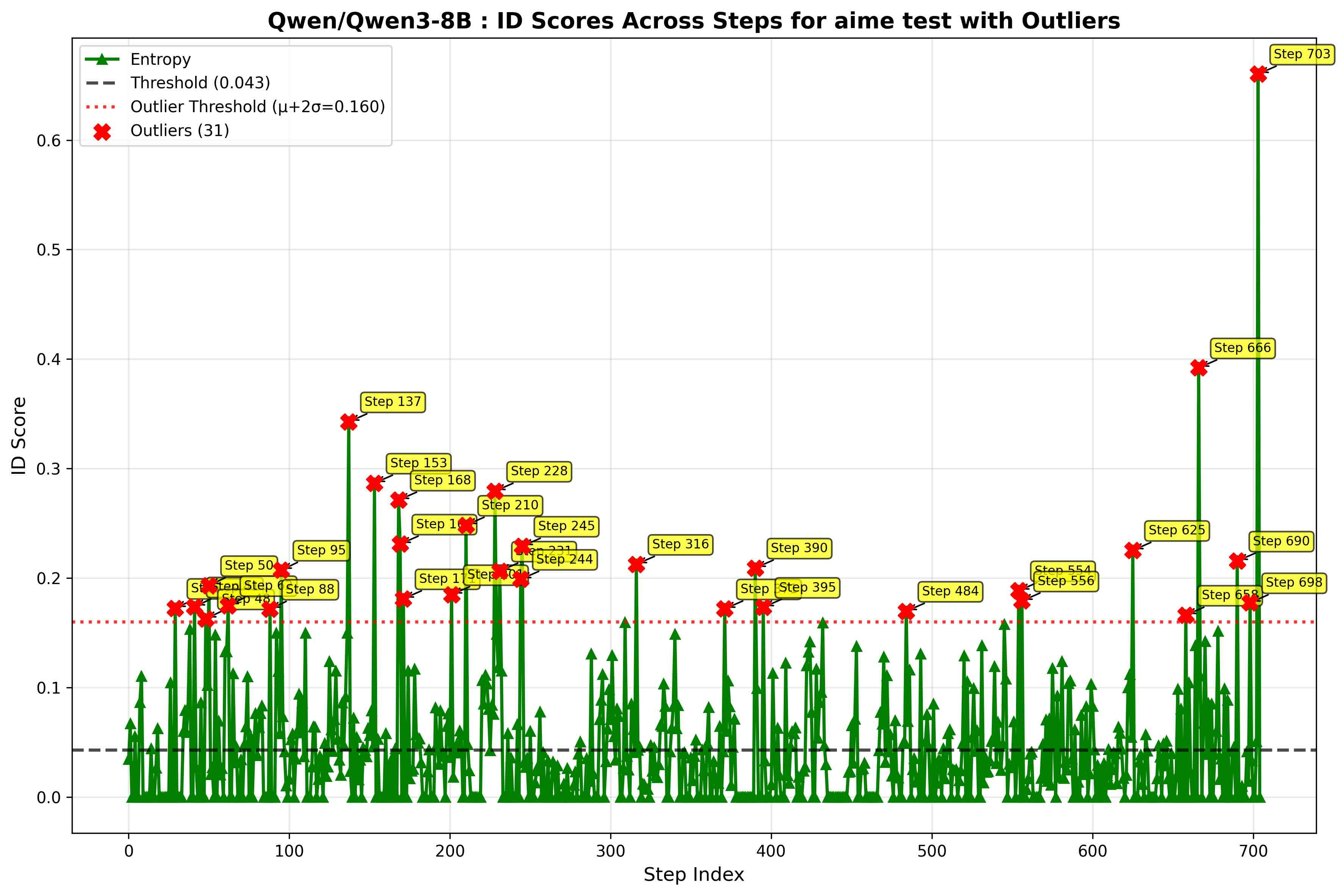}
\vspace{-0.1in}
\caption{Q11 — Correct Trace Visualization}
\vspace{-0.1in}
\label{fig:q11_correct_vis}
\end{figure*}

\begin{figure*}[t]
\centering
\includegraphics[width=\textwidth]{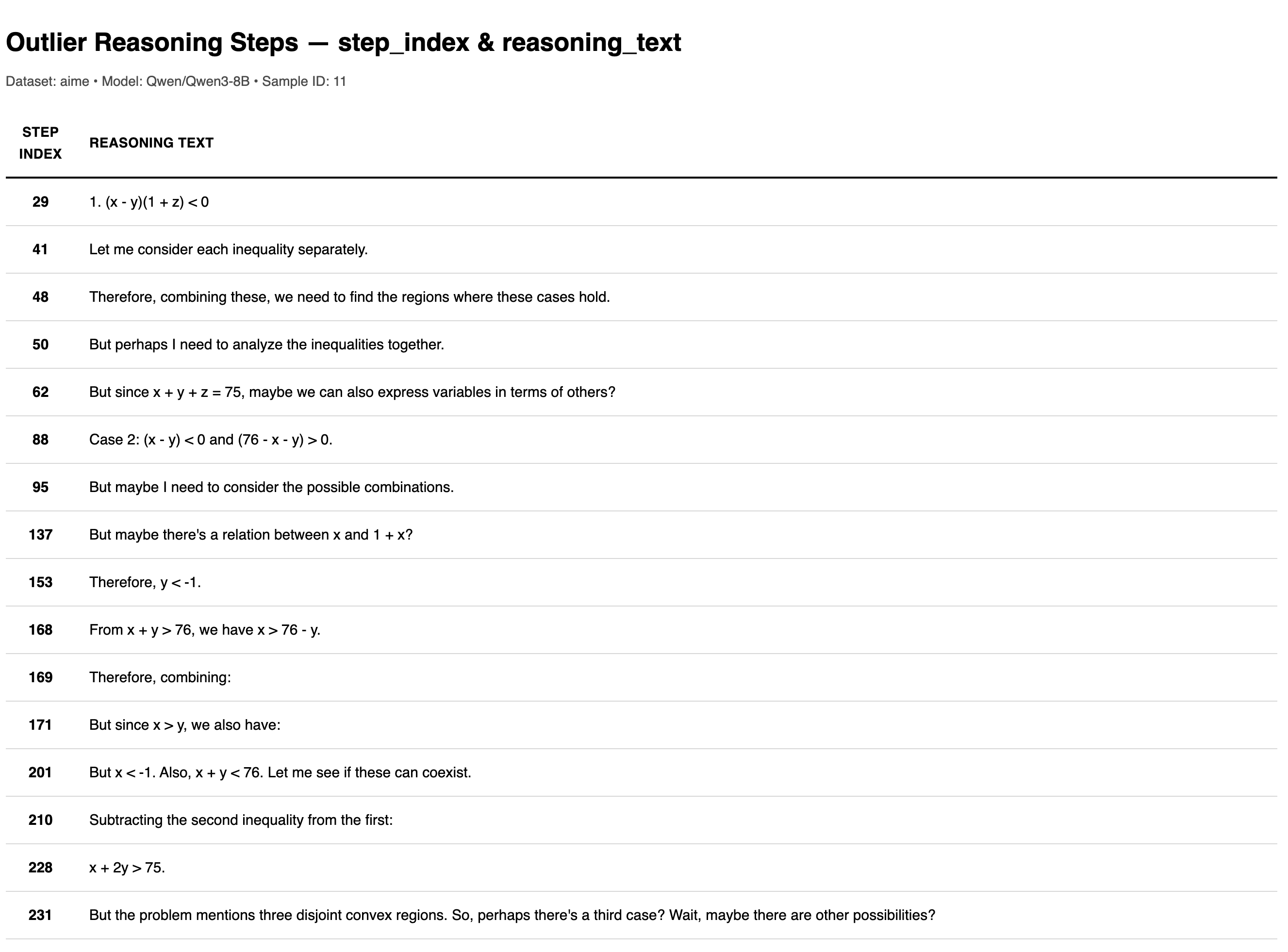}
\vspace{-0.1in}
\caption{Q11 — Correct Trace Text}
\vspace{-0.1in}
\label{fig:q11_correct_text}
\end{figure*}

\begin{figure*}[t]
\centering
\includegraphics[width=\textwidth]{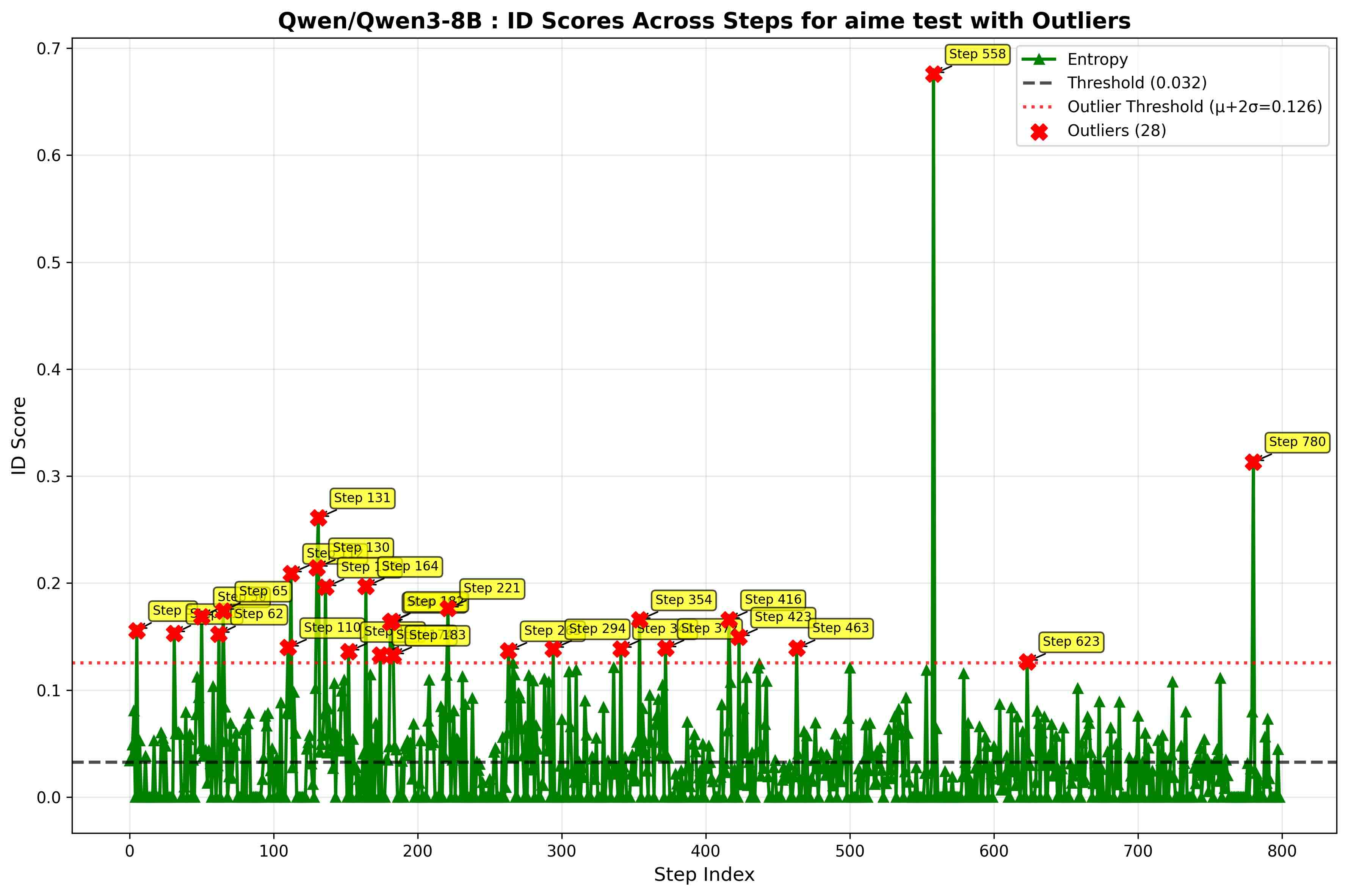}
\vspace{-0.1in}
\caption{Q11 — Incorrect Trace Visualization}
\vspace{0.1in}
\label{fig:q11_incorrect_vis}
\end{figure*}

\begin{figure*}[t]
\centering
\includegraphics[width=\textwidth]{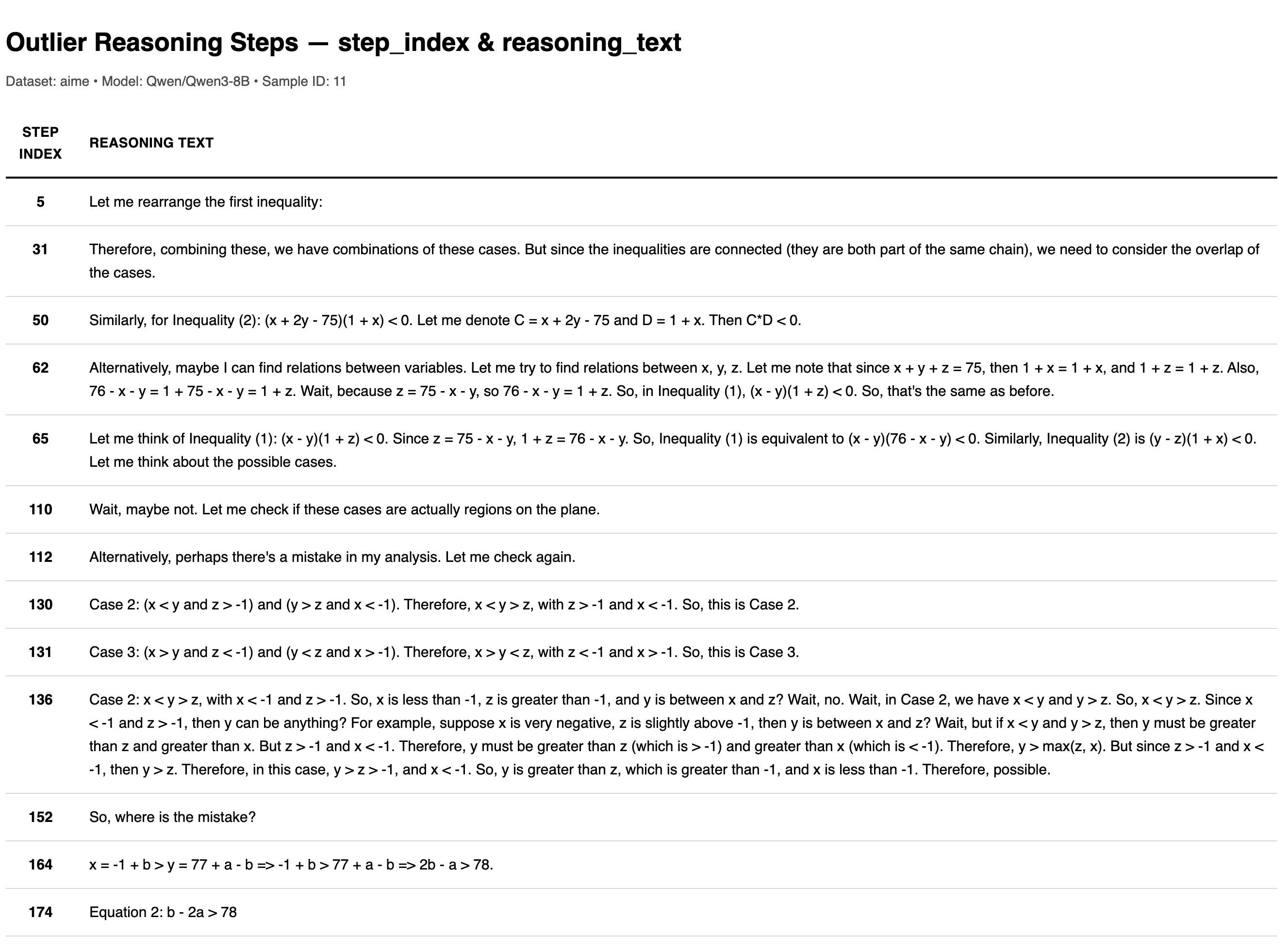}
\vspace{-0.1in}
\caption{Q11 — Incorrect Trace Text}
\vspace{-0.1in}
\label{fig:q11_incorrect_text}
\end{figure*}

\begin{figure*}[t]
\centering
\includegraphics[width=\textwidth]{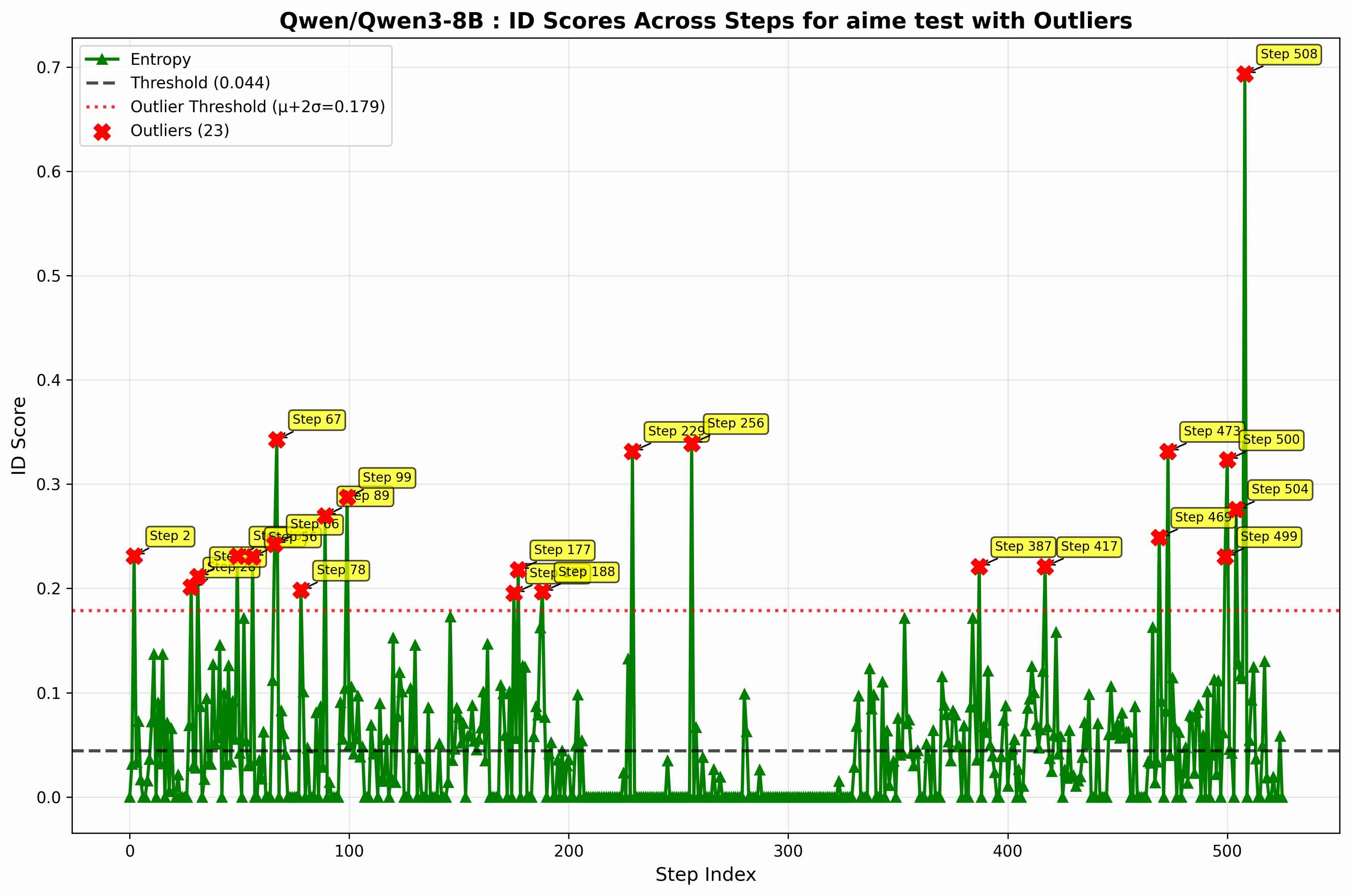}
\vspace{-0.1in}
\caption{Q17 — Correct Trace Visualization}
\vspace{0.1in}
\label{fig:q17_correct_vis}
\end{figure*}

\begin{figure*}[t]
\centering
\includegraphics[width=\textwidth]{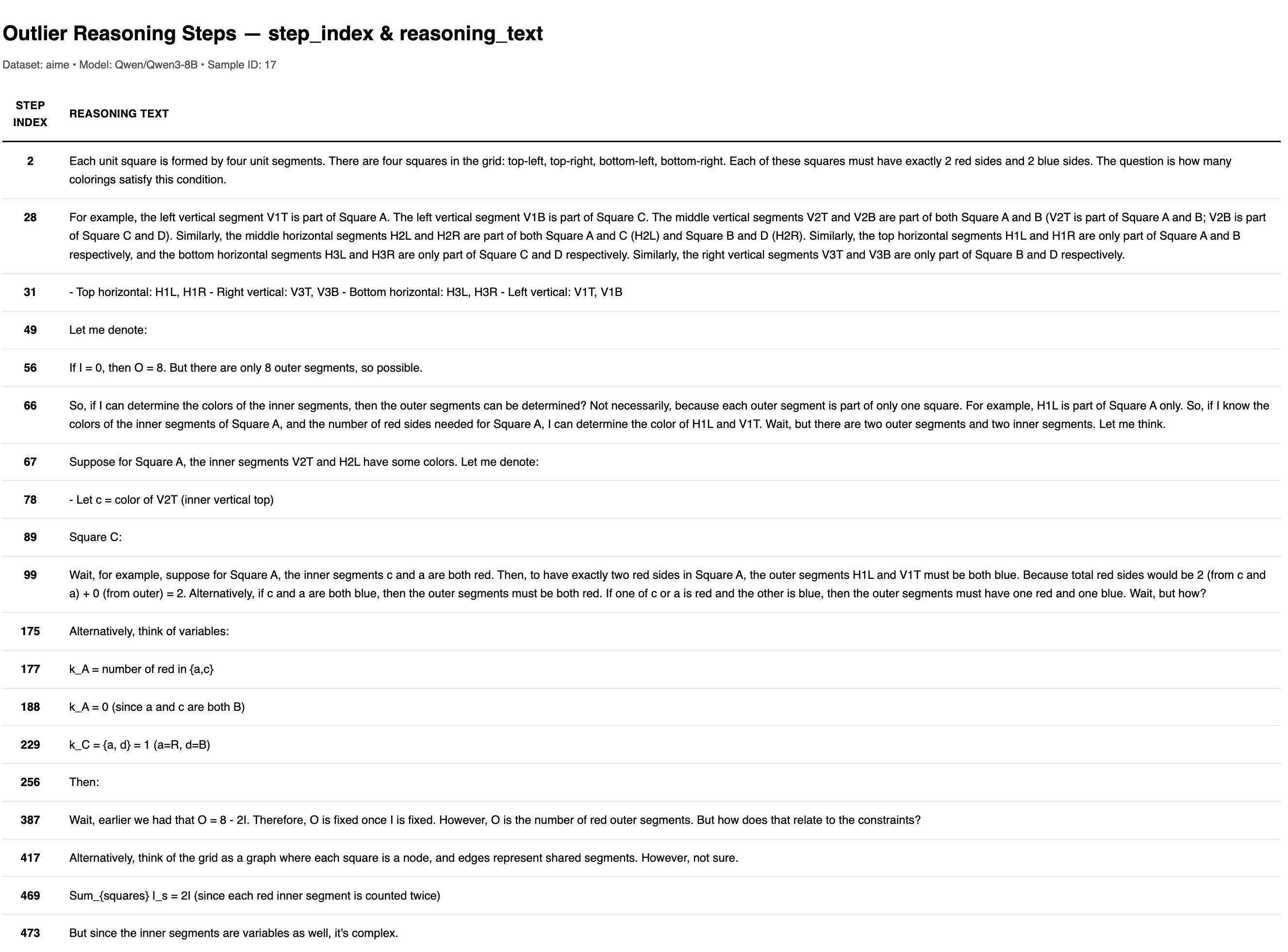}
\vspace{-0.1in}
\caption{Q17 — Correct Trace Text}
\vspace{-0.1in}
\label{fig:q17_correct_text}
\end{figure*}

\begin{figure*}[t]
\centering
\includegraphics[width=\textwidth]{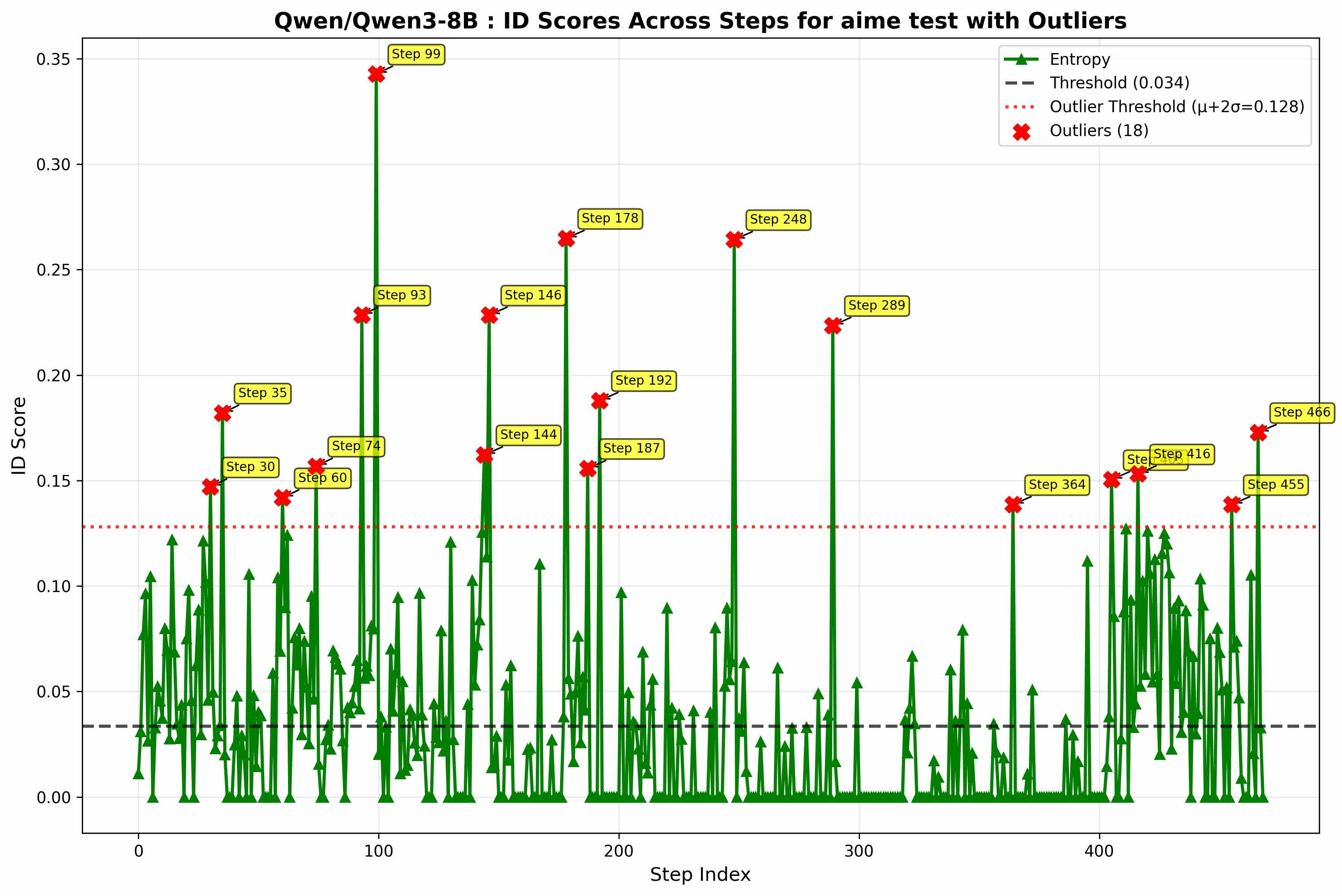}
\vspace{-0.1in}
\caption{Q17 — Incorrect Trace Visualization}
\vspace{-0.1in}
\label{fig:q17_incorrect_vis}
\end{figure*}

\begin{figure*}[t]
\centering
\includegraphics[width=\textwidth]{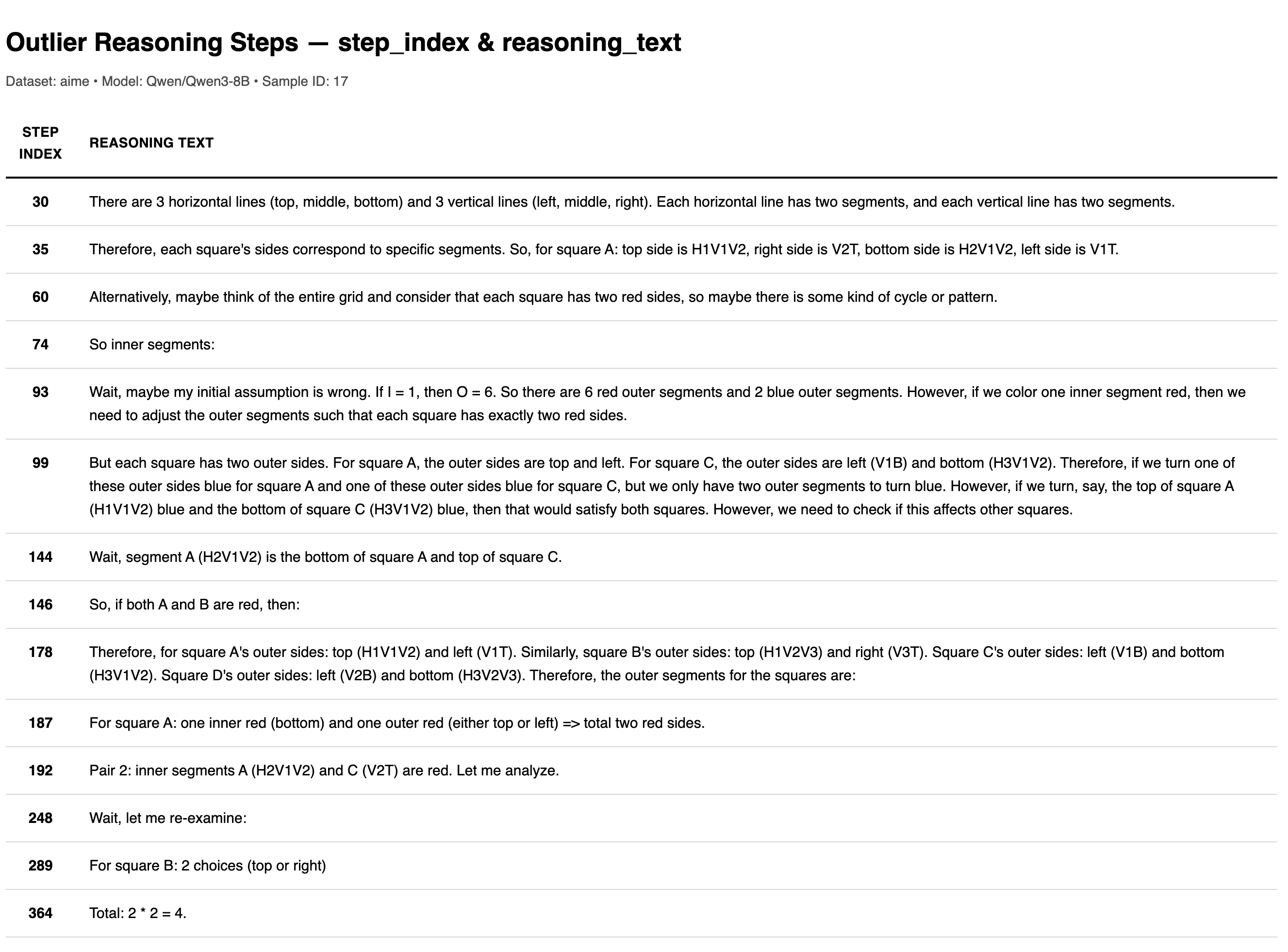}
\vspace{-0.1in}
\caption{Q17 — Incorrect Trace Text}
\vspace{-0.1in}
\label{fig:q17_incorrect_text}
\end{figure*}

Figure~\ref{fig:q6_correct_vis} to Figure~\ref{fig:q17_incorrect_text} are some examples of the case studies performed. We conduct trace-level analysis of reasoning steps of single instances of LLM response to see if we can significantly find patterns only at the token level by looking at entropy peaks where we define peaks as stpes above $u +2\sigma$ where $u$ and $\sigma$ are, respectively, average and standard deviations of the entropy value of individual steps. We use \texttt{Qwen3-8B} on AIME2025 dataset. For both correct and wrong reasoning traces, we (1) identify outlier steps on reasoning traces, and (2) identify their corresponding token-level steps to see if there are a meaningful differences between correct and incorrect traces. We also look at the number of total number of steps of both correct and wrong reasoning traces.

We identify that only qualitively identifying the reasoning traces is not enough to effectively discriminate high-and-low quality reasoning. To be specific, (1) entropy levels in both traces can be quite varying, (2) number of transition words \textit{(\ie But, Alternatively, Wait)} may appear more at correct traces contrary to our intuition, and (3) the number of steps may be shorter and concise at wrong reasoning traces.

\section{Sensitivity to Outlier Threshold $\tau$}
We initially chose $\tau$ to be 2 and 3 as mentioned in our appendix, and reported only $\tau$ = 3 as an intuitive "moderate outlier" criterion similar to z-test to avoid expensive hyperparameter tuning. However, to further explore the sensitivity of the results with with respect to the size of $\tau$ that captures local uniformity, we also ran a broader sweep with $\tau$ $\in$ \{2,3,4,5\} in Table~\ref{tab:ksweep_seed42}. The results show that increasing $\tau$ makes the disruption criterion stricter as fewer changes counted as disruptions, but the overall conclusions are robust: performance trends and the relative behavior of Loc. uni vs. Loc. non-uni do not hinge on a specific choice of k. Therefore, we justify that our $\tau$ is a statistical thresholding constant, and not a tuned hyperparameter.

\section{Implementation Details}\label{app:impl}

\subsection{Hyperparameters and GPUs.}

We use 2 x H100 GPUs for our main results on Qwen3-8B (thinking mode) and Deepseek-Distill-Qwen-7B, and 4 x RTX A6000 GPUs for all others. Temperature is 0.6, top-p 0.95, and top-k 20, as stated in \citep{yang2025qwen3technicalreport} and \citep{deepseekai2025deepseekr1incentivizingreasoningcapability}.

\subsection{Benchmarks}\label{app:bench}

\noindent\textbf{AIME 2025.}\footnote{\url{https://huggingface.co/datasets/opencompass/AIME2025}} The American Invitational Mathematics Examination (AIME) is a prestigious US high school math contest consisting of challenging integer-answer quetions. The AIME 2025 benchmark uses problems from the 2025 contests to evaluate a LLM's mathematical reasoning by requiring a single correct integer answer. The test set used in our analysis contains of 30 questions.

\noindent \textbf{BRUMO 2025.}\footnote{\url{https://huggingface.co/datasets/MathArena/brumo_2025}} The Brown University Math Olympiad (BRUMO) is a mathematics competition for students. The BRUMO 2025 benchmark is built from the problems of the 2025 BRUMO competition, and as part of the MathArena \citep{balunović2025matharenaevaluatingllmsuncontaminated} benchmark suite, it consists of 30 final-answer problems. Each requires a unique numeric or closed-form solution. Unlike proof-based contests, BRUMO problems are evaluated on answer correctness.

\noindent \textbf{HMMT 2025.}\footnote{\url{https://huggingface.co/datasets/MathArena/hmmt_feb_2025}} The Harvard-MIT Mathematics Tournament (HMMT) is a renowned competition featuring diverse problems in algebra, geometry, combinatorics, and number theory. The HMMT 2025 benchmark uses newly releassed problems from the February 2025 tournament, providing a broader variety of tasks than AIME. The set used in our analysis contains of 30 questions.

\noindent \textbf{MinervaMath.}\footnote{\url{https://huggingface.co/datasets/math-ai/minervamath}} The Minerva Math benchamrk consists of advanced quantitative problems sourced from university-level STEM courses, including physics, chemistry, and higher mathematics. The set used in our analysis contains of 272 questions.


\noindent \textbf{GPQA-Diamond.}\footnote{\url{https://huggingface.co/datasets/Idavidrein/gpqa}} GPQA-Diamond \citep{rein2023gpqagraduatelevelgoogleproofqa} is the hardest subset of the Graduate-Level Google-Proof Q\&A benchmark, consisting of 198 multiple-choice questions in biology, chemistry, and physics. These questions are crafted by domain experts and validated by multiple PhD-level validators to ensure clarity and high difficulty. The Google-proof design means that even with web search, solving them requires deep reasoning rather than lookup.

\noindent \textbf{LSAT-AR.}\footnote{\url{https://huggingface.co/datasets/allenai/lsat-ar}} The LSAT Analytical Reasoning (LSAT-AR) benchmark, also known as \emph{logic games}, evaluates a model’s ability to perform structured symbolic reasoning under explicit constraints. Each problem presents a set of entities together with a collection of rules, and requires deducing valid arrangements or assignments that satisfy all constraints. LSAT-AR problems are designed to test combinatorial reasoning, constraint satisfaction, and systematic deduction rather than factual knowledge. The benchmark is widely used to assess logical planning and rule-based reasoning abilities in language models.

\noindent \textbf{LSAT-LR.}\footnote{\url{https://huggingface.co/datasets/allenai/lsat-lr}} The LSAT Logical Reasoning (LSAT-LR) benchmark consists of short natural-language arguments followed by multiple-choice questions that probe logical validity, assumptions, and implications. Unlike LSAT-AR, LSAT-LR focuses on informal logical reasoning, including identifying flaws, strengthening or weakening arguments, and drawing valid conclusions from given premises. This benchmark evaluates a model’s ability to understand argument structure, implicit assumptions, and causal or logical relationships expressed in natural language.

\subsection{Baseline Details}\label{app:baselines}
We re-implemented all logic using vLLM, unlike some of the codes initially released. Our main baselines are selection methods that leverage LLMs' internal signals.

\noindent \textbf{Self-Certainty.} This is the implementation of \citet{kang2025scalablebestofnselectionlarge}, where it evaluates LLMs' reasoning by introducing a self-certainty, a confidence score assigned at each reasoning step. Self-certainty captures whether the model is confident in its logical steps, and uses Borda Voting to improve answer selection. Borda Voting ranks outputs by aggregared confidence rather than relying on simple majority voting used in Self-Consistency ~\citep{wang2023selfconsistencyimproveschainthought}.

\noindent \textbf{Highest Confidence.} This is a naive implementation that selects the path with the highest oeverall token confidence in the reasoning trace. Similar methods have been introduced in \citep{jang2025selftraininglargelanguagemodels}, where paths for training are selected based on the traces with the highest confidence.

\noindent \textbf{Lowest Entropy.} This is a naive implementation that selects the path with the lowest overall token entropy in the reasoning trace, driven from the idea that entropy itself is a measurement of uncertainty.

\section{Experiment Details}

\paragraph{Details of the main experiment}\label{app:main}
Details of all the experiments carried out on the three seeds (42, 1234, 2025) are in Table~\ref{tab:mainresultsseed-ds},~\ref{tab:mainresultssee-qwen}, and~\ref{tab:main-results-llama}.

\paragraph{Details of model size scaling experiment}\label{app:model-size}

Details of all the experiments carried out on the three seeds (42, 1234, 2025) are in Table~\ref{tab:scalemodelseed}.

\paragraph{Details of sample size scaling experiment}\label{app:sample-scale}

Details of all the experiments carried out on the three seeds (42, 1234, 2025) are in Table~\ref{tab:sample-scale}.

\section{Additional Analyses}
\subsection{Complementarity with majority voting}
\label{subsec:complementarity}
Our primary analysis focuses on \emph{trace-level selection}, where a single reasoning trajectory is chosen based on internal structural signals such as information-flow uniformity. In contrast, \emph{answer-level aggregation} methods—most notably majority voting—operate by comparing multiple independently generated responses and selecting answers based on cross-sample agreement. These two approaches therefore act at different granularities: UID evaluates how information evolves \emph{within} a single trace, whereas aggregation methods exploit consensus \emph{across} traces.

To clarify their relationship, we compare UID-based selectors with majority voting under \emph{matched sampling budgets}, and additionally evaluate combinations of UID with multi-sample aggregation in an extended analysis. In particular, we adopt \emph{Borda voting} as a lightweight aggregation mechanism over trace-level structural signals. Given $N$ sampled traces indexed by $i \in \{1,\dots,N\}$, each trace is assigned a rank $r_i$ according to a chosen criterion and converted into a weight
\[
w_i = (N - r_i + 1)^p,
\]
with exponent $p=0.5$ and $N = 3$ unless otherwise noted. The selected trace is given by $i^* = \arg\max_i w_i$.

We consider two Borda-based variants. \emph{(i) UID-based Borda voting} ranks traces in descending order of their UID scores, evaluating whether information-flow uniformity within a trace is predictive of correctness. \emph{(ii) Spikes/Falls-based Borda voting} ranks traces in ascending order of entropy irregularity, measured as the number of entropy spikes and falls, to examine whether local disruptions in entropy smoothness correlate with performance.

Table~\ref{tab:majority_voting} reports results across three backbone models. We find that UID-based selection can be competitive with majority voting and often complementary when combined with lightweight aggregation schemes such as Borda voting, consistent with~\citep{kang2025scalablebestofnselectionlarge}.

\subsection{Qualitative analyses on reasoning traces with low and high UID scores}\label{app:quali-analysis}
\begin{figure*}[t]
\centering
\includegraphics[width=\textwidth]{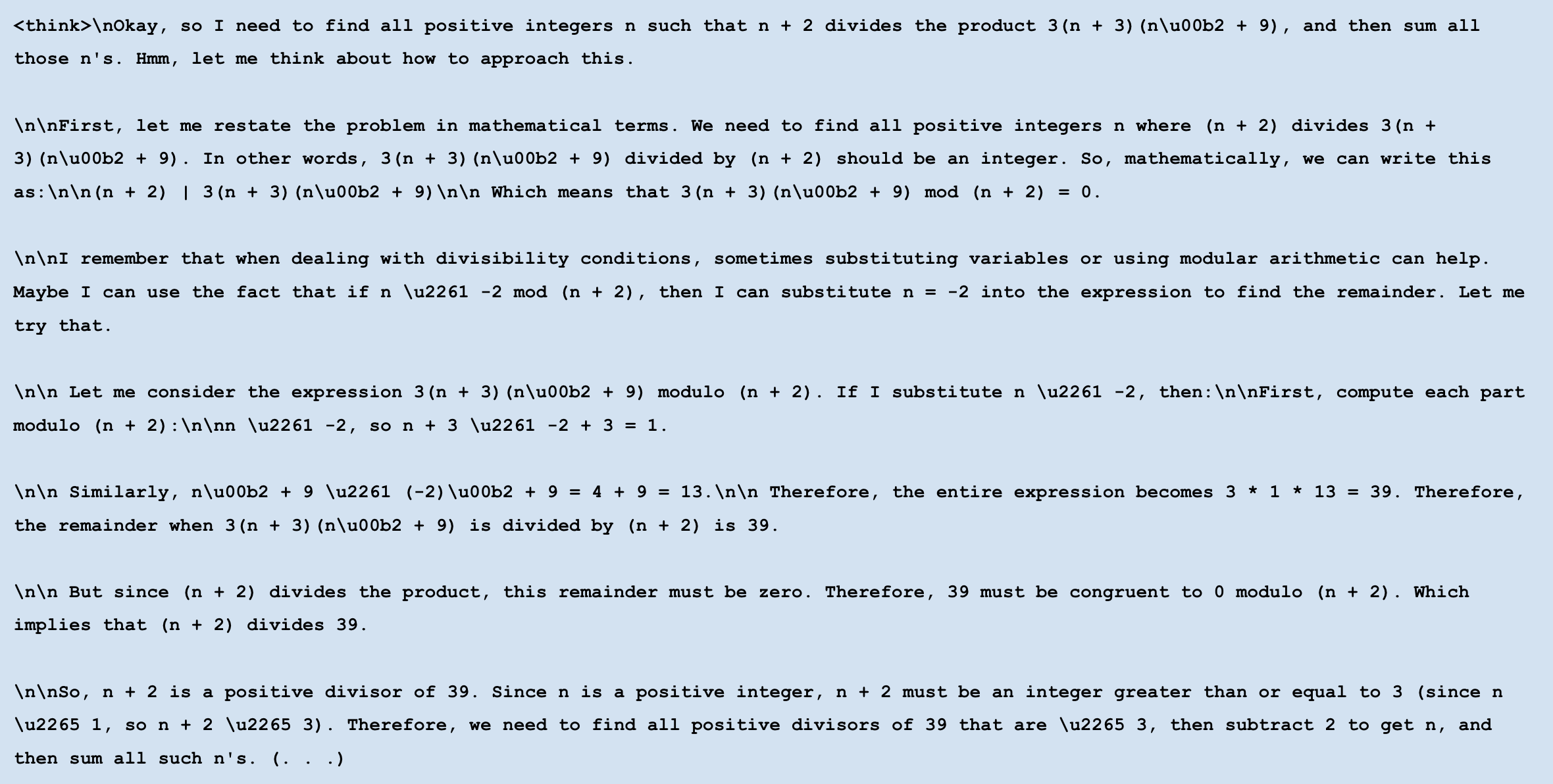}
\vspace{-0.1in}
\caption{Reasoning trace with the highest variance score, with rich reasoning. \texttt{math\_equal} is True}
\vspace{-0.1in}
\label{fig:highest-variance}
\end{figure*}

\begin{figure*}[t]
\centering
\includegraphics[width=\textwidth]{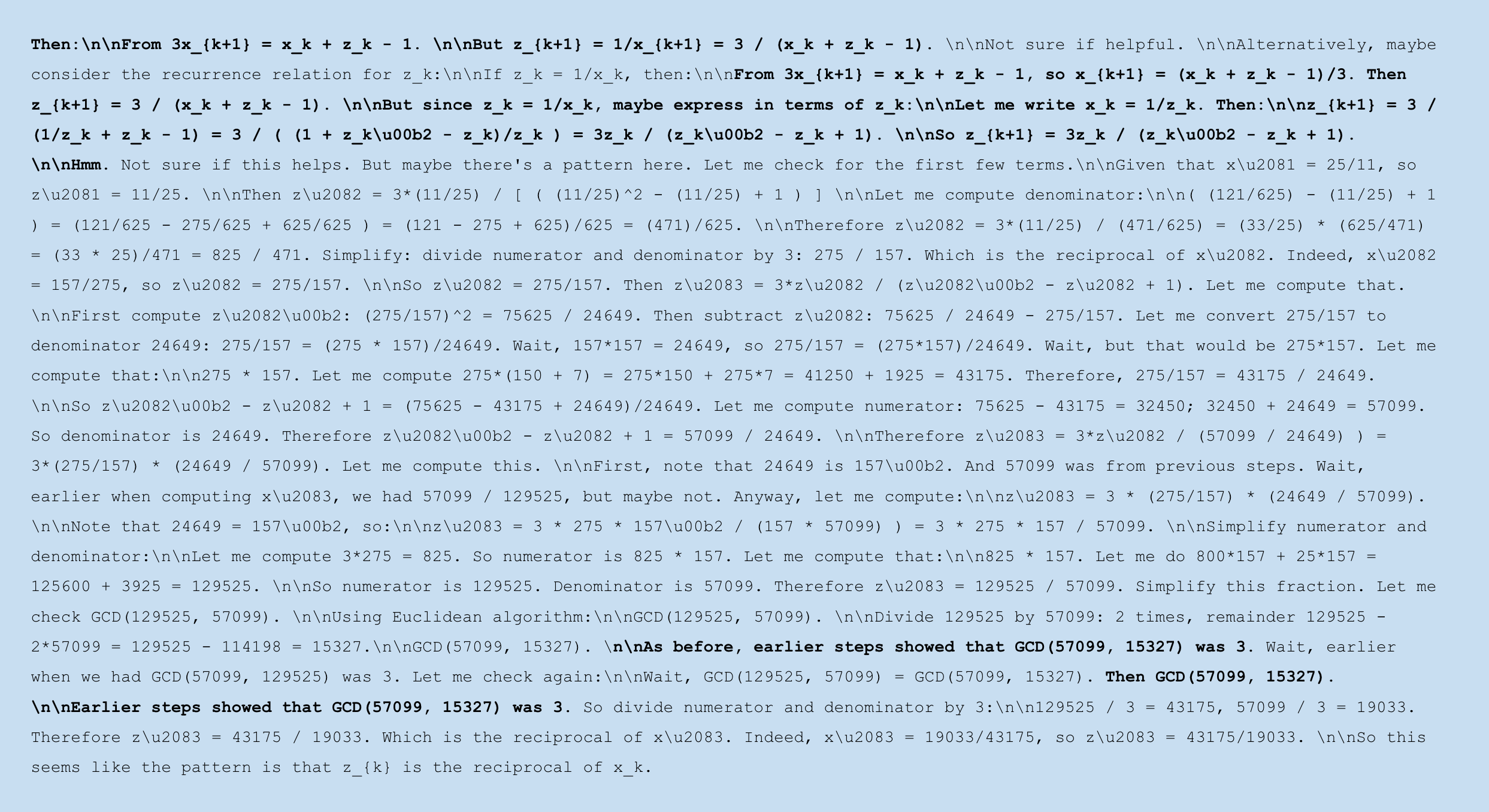}
\vspace{-0.1in}
\caption{Reasoning trace with the lowest variance score, showing that the model is stuck. \texttt{math\_equal} is False}
\vspace{-0.1in}
\label{fig:lowset-variance}
\end{figure*}

We qualitatively analyze reasoning traces with the highest and lowest UID scores across the dataset (Figures~\ref{fig:highest-variance} and \ref{fig:lowset-variance}). Traces that are globally non-uniform exhibit the characteristic ebb and flow of high-quality reasoning, reflecting meaningful shifts in information density. In contrast, low-variance traces reveal a failure mode of superficially uniform but uninformative expansion.

\begin{figure*}[t]
\centering
\includegraphics[width=\textwidth]{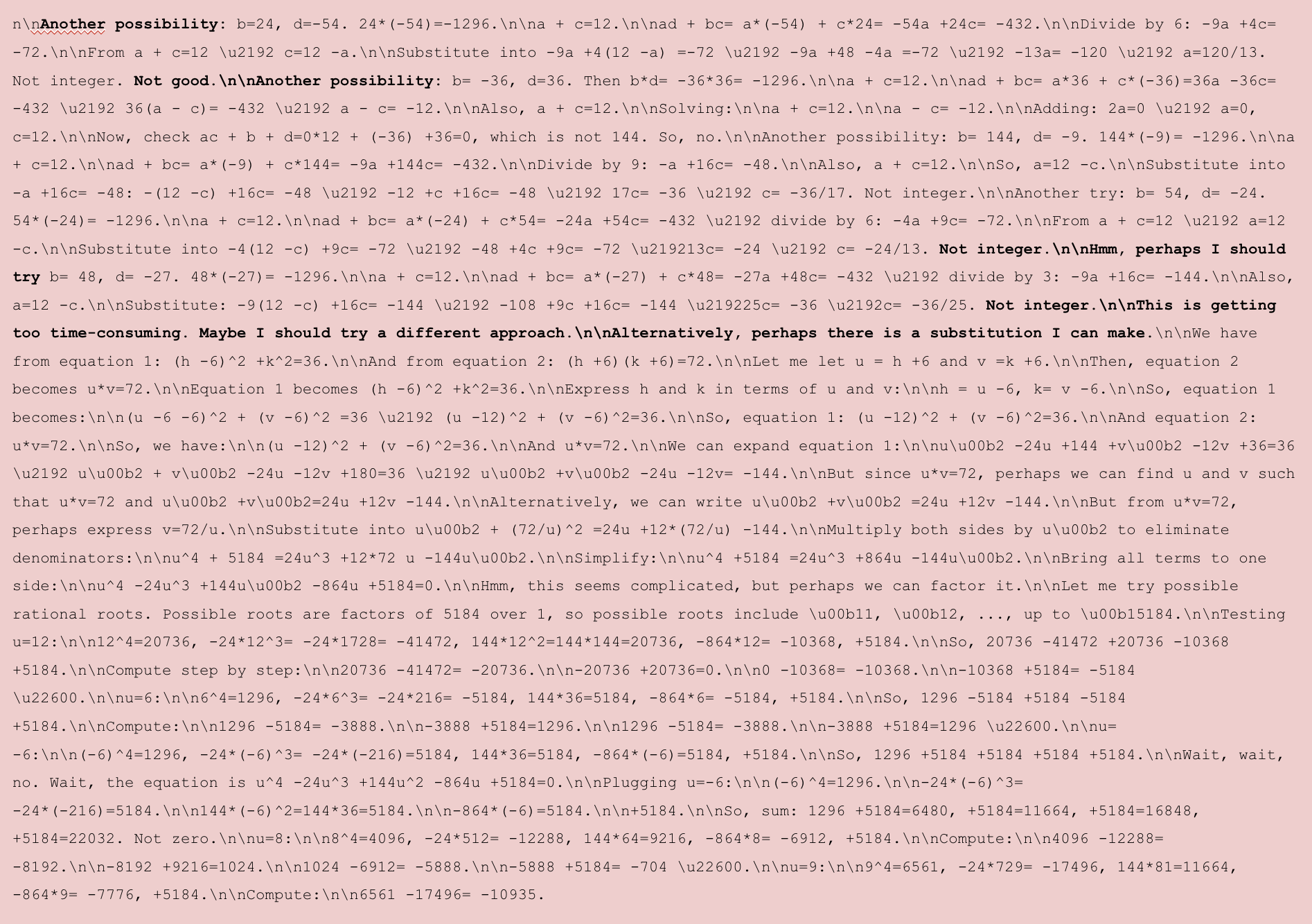}
\vspace{-0.1in}
\caption{Reasoning trace with the highest number of spikes and falls, with a lot of redundancy and confusions. \texttt{math\_equal} is True}
\vspace{-0.1in}
\label{fig:highest-spikes-falls}
\end{figure*}

\begin{figure*}[t]
\centering
\includegraphics[width=\textwidth]{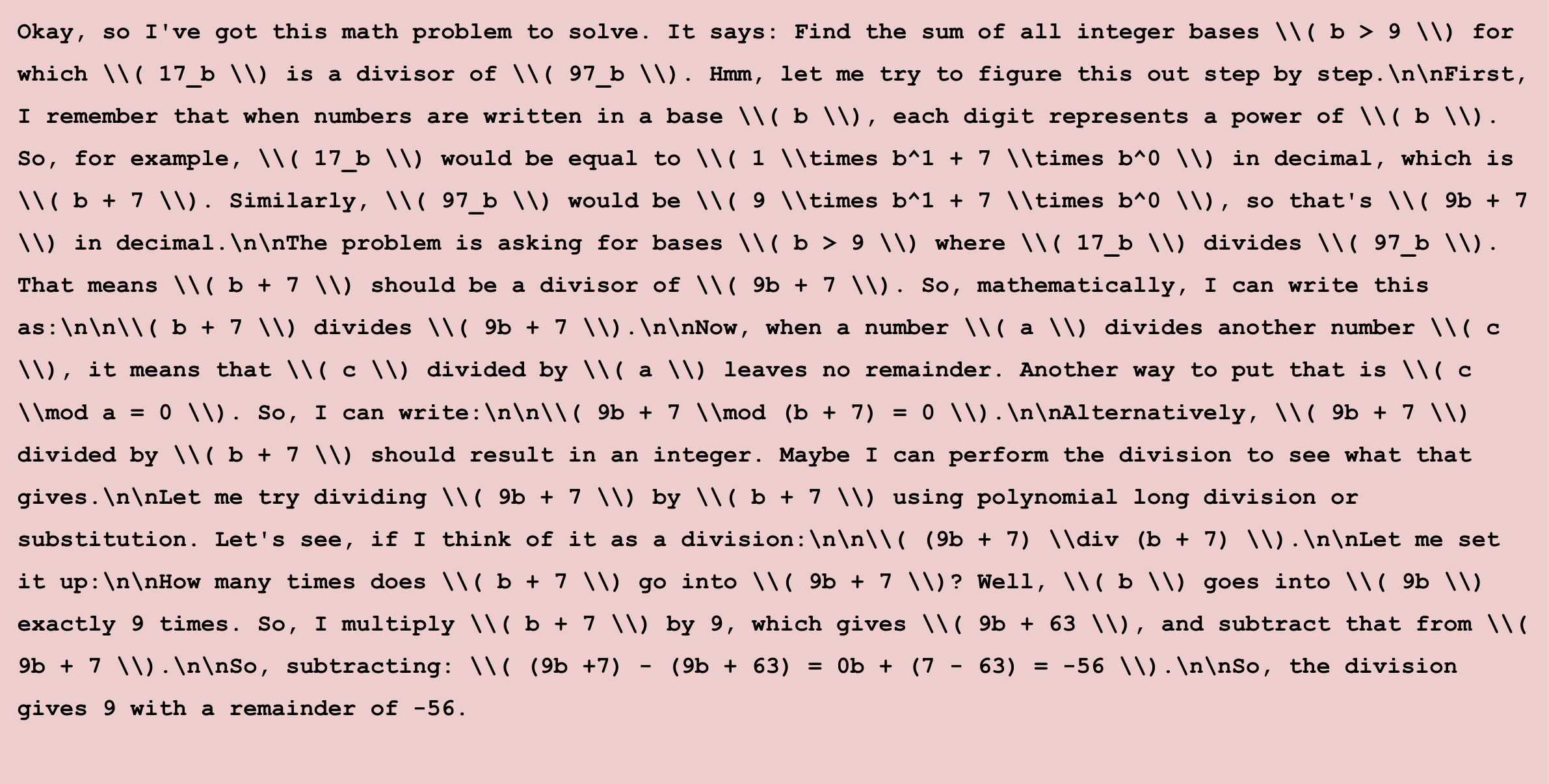}
\vspace{-0.1in}
\caption{Reasoning trace with the lowest number of spikes and falls, with simple and concise reasoning. \texttt{math\_equal} is True}
\vspace{-0.1in}
\label{fig:lowset-spikes-falls}
\end{figure*}

At a local scale, Figure~\ref{fig:highest-spikes-falls} illustrates a reasoning trace that is locally non-uniform yet stuck and repetitive, despite arriving at the correct answer. Conversely, locally uniform traces (Figure~\ref{fig:lowset-spikes-falls}) avoid such degenerative repetition and maintain coherent progress.

\section{Prompts for Communicative vs. Naive Prompting}\label{app:prompts}
Details of the prompts used for comparing human communication and LLM reasoning are attached in Table~\ref{tab:prompts}. We evaluated it on AIME2025 benchmark, sampled by $n = 3$.

\section{Usage of AI Assistants}
In preparing this work, we used AI-based writing assistants to improve sentence structure, correct grammatical errors, and enhance overall readability. These tools were employed solely for language refinement and did not contribute to the development of technical content, research methodology, or experimental analysis. All scientific ideas, results, and conclusions presented in the paper were conceived and authored entirely by the researchers. Use of AI assistance was restricted to editorial purposes and did not affect the originality or intellectual contributions of the work.

\FloatBarrier

\clearpage

\input{Tables/MainResultsSeed_ds}
\input{Tables/MainResultsSeed_qwen3}
\input{Tables/MainReesultsSeed_llama}
\clearpage
\input{Tables/ScaleModelSeed}
\input{Tables/ScaleSampleSeed}
\clearpage
\input{Tables/MajorityVoting}

\input{Tables/Prompts}

\clearpage

\clearpage
\input{Tables/K_Sweep_Seed42}



%% file: Tables/APP_table6.tex
\begin{table*}[t]
    \centering
    \caption{Fixed window segmentation (2048 tokens)}
    \label{tab:fixed_window}
    
    \begin{adjustbox}{width=\textwidth}
    \begin{tabular}{l|ccc|ccc|ccc}
        \toprule
        & \multicolumn{3}{c}{\texttt{Qwen3-8B}} 
        & \multicolumn{3}{c}{\texttt{DS-R1-Distill-Qwen-7B}} 
        & \multicolumn{3}{c}{\texttt{DS-R1-Distill-Llama-8B}} \\
        \cmidrule(lr){2-4} \cmidrule(lr){5-7} \cmidrule(lr){8-10}
        Setting & AIME & HMMT & BRUMO 
        & AIME & HMMT & BRUMO 
        & AIME & HMMT & BRUMO \\
        \midrule
        
        Loc. non-uni ($3\sigma$)
        & \textbf{0.77} & 0.47 & 0.70
        & 0.40 & 0.23 & \textbf{0.57}
        & \textbf{0.40} & 0.14 & 0.33 \\

        \rowcolor{lightblue}
        Loc. uni ($3\sigma$)
        & \textbf{0.77} & \textbf{0.47} & \textbf{0.70}
        & 0.40 & 0.23 & \textbf{0.57}
        & \textbf{0.40} & 0.14 & 0.33 \\

        \rowcolor{lightblue}
        Glob. non-uni (var)
        & 0.67 & \textbf{0.43} & 0.63
        & 0.40 & \textbf{0.30} & \textbf{0.60}
        & \textbf{0.40} & \textbf{0.27} & \textbf{0.43} \\

        Glob. uni (var)
        & 0.63 & 0.33 & \textbf{0.67}
        & 0.27 & 0.20 & 0.37
        & 0.27 & 0.20 & 0.33 \\
        
        \bottomrule
    \end{tabular}
    \end{adjustbox}
\end{table*}

%% file: Tables/APP_table7.tex
\begin{table*}[t]
    \centering
    \caption{Semantic Segmentation}
    \label{tab:semantic_seg}
    
    \begin{adjustbox}{width=\textwidth}
    \begin{tabular}{l|ccc|ccc|ccc}
        \toprule
        & \multicolumn{3}{c}{\texttt{Qwen3-8B}} 
        & \multicolumn{3}{c}{\texttt{DS-R1-Distill-Qwen-7B}} 
        & \multicolumn{3}{c}{\texttt{DS-R1-Distill-Llama-8B}} \\
        \cmidrule(lr){2-4} \cmidrule(lr){5-7} \cmidrule(lr){8-10}
        Setting & AIME & HMMT & BRUMO 
        & AIME & HMMT & BRUMO 
        & AIME & HMMT & BRUMO \\
        \midrule
        
        Loc. non-uni ($3\sigma$)
        & \textbf{0.70} & 0.33 & 0.60
        & 0.37 & 0.17 & 0.40
        & 0.20 & 0.20 & \textbf{0.43} \\

        \rowcolor{lightblue}
        Loc. uni ($3\sigma$)
        & 0.63 & \textbf{0.47} & \textbf{0.67}
        & \textbf{0.47} & \textbf{0.27} & \textbf{0.50}
        & \textbf{0.43} & 0.23 & 0.40 \\

        \rowcolor{lightblue}
        Glob. non-uni (var)
        & 0.60 & \textbf{0.43} & \textbf{0.67}
        & 0.27 & 0.20 & \textbf{0.53}
        & 0.30 & \textbf{0.27} & 0.40 \\

        Glob. uni (var)
        & 0.57 & 0.33 & \textbf{0.67}
        & 0.43 & \textbf{0.27} & 0.47
        & 0.37 & 0.20 & 0.40 \\
        
        \bottomrule
    \end{tabular}
    \end{adjustbox}
\end{table*}

%% file: Tables/MainResultsSeed_ds.tex
\begin{table*}[t]
\caption{\textbf{Main results.} Results of \texttt{Deepseek-R1-Distill-Qwen-7b} on four math benchmarks (AIME2025, BRUNO2025, HMMT2025, and MinervaMath). Performance is reported across seeds 42, 1234, and 2025, with averages and standard deviations (Avg~$\pm$~Std).}
\label{tab:mainresultsseed-ds}
\centering

\resizebox{0.98\textwidth}{!}{%
\begin{tabular}{@{}l|cccc|cccc@{}}
\toprule
\multirow{2}{*}{Methods ($\downarrow$)} &
\multicolumn{4}{c|}{AIME2025} &
\multicolumn{4}{c}{BRUMO2025} \\ \cmidrule(lr){2-5}\cmidrule(lr){6-9}
 & 42 & 1234 & 2025 & Avg $\pm$ Std
 & 42 & 1234 & 2025 & Avg $\pm$ Std \\ \midrule

Mean Acc.
& 0.41 & 0.39 & 0.41 & 0.40 $\pm$ {\footnotesize 0.02}
& 0.55 & 0.53 & 0.53 & 0.54 $\pm$ {\footnotesize 0.01} \\

Self-Cert.
& 0.50 & 0.43 & 0.50 & 0.48 $\pm$ {\footnotesize 0.04}
& 0.50 & 0.57 & 0.50 & 0.52 $\pm$ {\footnotesize 0.04} \\

High Conf.
& 0.50 & 0.43 & 0.50 & 0.48 $\pm$ {\footnotesize 0.04}
& 0.57 & 0.53 & 0.47 & 0.52 $\pm$ {\footnotesize 0.05} \\

Low Ent.
& 0.50 & 0.43 & 0.50 & 0.48 $\pm$ {\footnotesize 0.04}
& 0.60 & 0.57 & 0.50 & 0.56 $\pm$ {\footnotesize 0.05} \\

\midrule
UID Metrics ($\downarrow$) \\
\midrule

Loc. non-uni (2$\sigma$)
& 0.27 & 0.27 & 0.27 & 0.27 $\pm$ {\footnotesize 0.00}
& 0.47 & 0.33 & 0.37 & 0.39 $\pm$ {\footnotesize 0.07} \\

\rowcolor{lightblue}
Loc. uni (2$\sigma$)
& 0.53 & 0.47 & 0.53 & 0.51 $\pm$ {\footnotesize 0.04}
& 0.57 & 0.57 & 0.57 & 0.57 $\pm$ {\footnotesize 0.00} \\

Loc. non-uni (3$\sigma$)
& 0.27 & 0.27 & 0.27 & 0.27 $\pm$ {\footnotesize 0.00}
& 0.47 & 0.33 & 0.37 & 0.39 $\pm$ {\footnotesize 0.07} \\

\rowcolor{lightblue}
Loc. uni (3$\sigma$)
& 0.57 & 0.47 & 0.57 & 0.53 $\pm$ {\footnotesize 0.06}
& 0.57 & 0.57 & 0.53 & 0.56 $\pm$ {\footnotesize 0.02} \\

\rowcolor{lightblue}
Glob. non-uni (var)
& 0.57 & 0.43 & 0.57 & 0.52 $\pm$ {\footnotesize 0.08}
& 0.60 & 0.63 & 0.70 & 0.64 $\pm$ {\footnotesize 0.05} \\

Glob. uni (var)
& 0.33 & 0.33 & 0.33 & 0.33 $\pm$ {\footnotesize 0.00}
& 0.47 & 0.43 & 0.43 & 0.44 $\pm$ {\footnotesize 0.02} \\

\bottomrule
\end{tabular}
}

\vspace{6pt}

\resizebox{0.98\textwidth}{!}{%
\begin{tabular}{@{}l|cccc|cccc@{}}
\toprule
\multirow{2}{*}{Methods ($\downarrow$)} &
\multicolumn{4}{c|}{HMMT2025} &
\multicolumn{4}{c}{MinervaMath} \\ \cmidrule(lr){2-5}\cmidrule(lr){6-9}
 & 42 & 1234 & 2025 & Avg $\pm$ Std
 & 42 & 1234 & 2025 & Avg $\pm$ Std \\ \midrule

Mean Acc.
& 0.24 & 0.23 & 0.23 & 0.24 $\pm$ {\footnotesize 0.00}
& 0.30 & 0.29 & 0.30 & 0.30 $\pm$ {\footnotesize 0.00} \\

Self-Cert.
& 0.30 & 0.27 & 0.27 & 0.28 $\pm$ {\footnotesize 0.02}
& 0.31 & 0.30 & 0.31 & 0.30 $\pm$ {\footnotesize 0.00} \\

High Conf.
& 0.27 & 0.27 & 0.27 & 0.27 $\pm$ {\footnotesize 0.00}
& 0.31 & 0.31 & 0.31 & 0.31 $\pm$ {\footnotesize 0.00} \\

Low Ent.
& 0.27 & 0.23 & 0.23 & 0.24 $\pm$ {\footnotesize 0.02}
& 0.30 & 0.31 & 0.30 & 0.30 $\pm$ {\footnotesize 0.00} \\

\midrule
UID Metrics ($\downarrow$) \\
\midrule

Loc. non-uni (2$\sigma$)
& 0.13 & 0.20 & 0.20 & 0.18 $\pm$ {\footnotesize 0.04}
& 0.27 & 0.27 & 0.27 & 0.27 $\pm$ {\footnotesize 0.00} \\

\rowcolor{lightblue}
Loc. uni (2$\sigma$)
& 0.30 & 0.30 & 0.30 & 0.30 $\pm$ {\footnotesize 0.00}
& 0.30 & 0.32 & 0.30 & 0.31 $\pm$ {\footnotesize 0.01} \\

Loc. non-uni (3$\sigma$)
& 0.13 & 0.20 & 0.20 & 0.18 $\pm$ {\footnotesize 0.04}
& 0.26 & 0.27 & 0.26 & 0.26 $\pm$ {\footnotesize 0.01} \\

\rowcolor{lightblue}
Loc. uni (3$\sigma$)
& 0.30 & 0.30 & 0.30 & 0.30 $\pm$ {\footnotesize 0.00}
& 0.31 & 0.32 & 0.31 & 0.31 $\pm$ {\footnotesize 0.01} \\

\rowcolor{lightblue}
Glob. non-uni (var)
& 0.30 & 0.23 & 0.23 & 0.26 $\pm$ {\footnotesize 0.04}
& 0.30 & 0.31 & 0.30 & 0.30 $\pm$ {\footnotesize 0.01} \\

Glob. uni (var)
& 0.20 & 0.13 & 0.13 & 0.16 $\pm$ {\footnotesize 0.04}
& 0.28 & 0.28 & 0.28 & 0.28 $\pm$ {\footnotesize 0.00} \\

\bottomrule
\end{tabular}
}

\end{table*}

%% file: Tables/MainResultsSeed_qwen3.tex
\begin{table*}[t]
\caption{\textbf{Main results.} Results of \texttt{Qwen3-8B} on four math benchmarks (AIME2025, BRUMO2025, HMMT2025, and MinervaMath). Performance is reported across seeds 42, 1234, and 2025, with averages and standard deviations (Avg~$\pm$~Std).}
\label{tab:mainresultssee-qwen}
\centering

\resizebox{0.98\textwidth}{!}{%
\begin{tabular}{@{}l|cccc|cccc@{}}
\toprule
\multirow{2}{*}{Methods ($\downarrow$)} &
\multicolumn{4}{c|}{AIME2025} &
\multicolumn{4}{c}{BRUMO2025} \\ \cmidrule(lr){2-5}\cmidrule(lr){6-9}
 & 42 & 1234 & 2025 & Avg $\pm$ Std
 & 42 & 1234 & 2025 & Avg $\pm$ Std \\ \midrule

Mean Acc.
& 0.66 & 0.66 & 0.68 & 0.67 $\pm$ {\footnotesize 0.01}
& 0.67 & 0.71 & 0.67 & 0.68 $\pm$ {\footnotesize 0.02} \\

Self-Cert.
& 0.63 & 0.63 & 0.63 & 0.63 $\pm$ {\footnotesize 0.00}
& 0.70 & 0.73 & 0.70 & 0.71 $\pm$ {\footnotesize 0.02} \\

High Conf.
& 0.60 & 0.60 & 0.60 & 0.60 $\pm$ {\footnotesize 0.00}
& 0.63 & 0.67 & 0.60 & 0.63 $\pm$ {\footnotesize 0.03} \\

Low Ent.
& 0.60 & 0.60 & 0.60 & 0.60 $\pm$ {\footnotesize 0.00}
& 0.63 & 0.70 & 0.60 & 0.64 $\pm$ {\footnotesize 0.04} \\

\midrule
UID Metrics ($\downarrow$) \\
\midrule

Loc. non-uni (2$\sigma$)
& 0.63 & 0.63 & 0.63 & 0.63 $\pm$ {\footnotesize 0.00}
& 0.53 & 0.70 & 0.67 & 0.63 $\pm$ {\footnotesize 0.07} \\

\rowcolor{lightblue}
Loc. uni (2$\sigma$)
& 0.67 & 0.67 & 0.73 & 0.69 $\pm$ {\footnotesize 0.03}
& 0.67 & 0.73 & 0.70 & 0.70 $\pm$ {\footnotesize 0.03} \\

Loc. non-uni (3$\sigma$)
& 0.63 & 0.63 & 0.63 & 0.63 $\pm$ {\footnotesize 0.00}
& 0.53 & 0.70 & 0.67 & 0.63 $\pm$ {\footnotesize 0.07} \\

\rowcolor{lightblue}
Loc. uni (3$\sigma$)
& 0.67 & 0.67 & 0.73 & 0.69 $\pm$ {\footnotesize 0.03}
& 0.67 & 0.73 & 0.70 & 0.70 $\pm$ {\footnotesize 0.03} \\

\rowcolor{lightblue}
Glob. non-uni (var)
& 0.70 & 0.70 & 0.70 & 0.70 $\pm$ {\footnotesize 0.00}
& 0.53 & 0.63 & 0.67 & 0.61 $\pm$ {\footnotesize 0.06} \\

Glob. uni (var)
& 0.67 & 0.67 & 0.63 & 0.66 $\pm$ {\footnotesize 0.02}
& 0.60 & 0.77 & 0.67 & 0.68 $\pm$ {\footnotesize 0.07} \\

\bottomrule
\end{tabular}
}

\vspace{6pt}

\resizebox{0.98\textwidth}{!}{%
\begin{tabular}{@{}l|cccc|cccc@{}}
\toprule
\multirow{2}{*}{Methods ($\downarrow$)} &
\multicolumn{4}{c|}{HMMT2025} &
\multicolumn{4}{c}{MinervaMath} \\ \cmidrule(lr){2-5}\cmidrule(lr){6-9}
 & 42 & 1234 & 2025 & Avg $\pm$ Std
 & 42 & 1234 & 2025 & Avg $\pm$ Std \\ \midrule

Mean Acc.
& 0.44 & 0.41 & 0.43 & 0.43 $\pm$ {\footnotesize 0.01}
& 0.34 & 0.34 & 0.33 & 0.34 $\pm$ {\footnotesize 0.01} \\

Self-Cert.
& 0.50 & 0.47 & 0.53 & 0.50 $\pm$ {\footnotesize 0.03}
& 0.34 & 0.35 & 0.34 & 0.34 $\pm$ {\footnotesize 0.01} \\

High Conf.
& 0.50 & 0.40 & 0.37 & 0.42 $\pm$ {\footnotesize 0.06}
& 0.33 & 0.34 & 0.33 & 0.33 $\pm$ {\footnotesize 0.01} \\

Low Ent.
& 0.50 & 0.40 & 0.40 & 0.43 $\pm$ {\footnotesize 0.05}
& 0.34 & 0.34 & 0.32 & 0.33 $\pm$ {\footnotesize 0.01} \\

\midrule
UID Metrics ($\downarrow$) \\
\midrule

Loc. non-uni (2$\sigma$)
& 0.43 & 0.37 & 0.40 & 0.40 $\pm$ {\footnotesize 0.03}
& 0.35 & 0.34 & 0.32 & 0.34 $\pm$ {\footnotesize 0.01} \\

\rowcolor{lightblue}
Loc. uni (2$\sigma$)
& 0.53 & 0.40 & 0.47 & 0.47 $\pm$ {\footnotesize 0.05}
& 0.34 & 0.34 & 0.33 & 0.34 $\pm$ {\footnotesize 0.01} \\

Loc. non-uni (3$\sigma$)
& 0.43 & 0.37 & 0.40 & 0.40 $\pm$ {\footnotesize 0.03}
& 0.35 & 0.34 & 0.32 & 0.34 $\pm$ {\footnotesize 0.01} \\

\rowcolor{lightblue}
Loc. uni (3$\sigma$)
& 0.53 & 0.43 & 0.47 & 0.48 $\pm$ {\footnotesize 0.04}
& 0.34 & 0.34 & 0.33 & 0.34 $\pm$ {\footnotesize 0.01} \\

\rowcolor{lightblue}
Glob. non-uni (var)
& 0.50 & 0.47 & 0.43 & 0.47 $\pm$ {\footnotesize 0.03}
& 0.35 & 0.33 & 0.32 & 0.33 $\pm$ {\footnotesize 0.01} \\

Glob. uni (var)
& 0.43 & 0.40 & 0.40 & 0.41 $\pm$ {\footnotesize 0.02}
& 0.35 & 0.35 & 0.31 & 0.34 $\pm$ {\footnotesize 0.02} \\

\bottomrule
\end{tabular}
}

\end{table*}

%% file: Tables/MainReesultsSeed_llama.tex
\begin{table*}[t]
\caption{\textbf{Main results.} Results of \texttt{Deepseek-R1-Distill-Llama-8b} on four math benchmarks (AIME2025, BRUMO2025, HMMT2025, and MinervaMath). Performance is reported across seeds 42, 1234, and 2025, with averages and standard deviations (Avg~$\pm$~Std).}
\label{tab:main-results-llama}
\centering

\resizebox{0.98\textwidth}{!}{%
\begin{tabular}{@{}l|cccc|cccc@{}}
\toprule
\multirow{2}{*}{Methods ($\downarrow$)} &
\multicolumn{4}{c|}{AIME2025} &
\multicolumn{4}{c}{BRUMO2025} \\ \cmidrule(lr){2-5}\cmidrule(lr){6-9}
 & 42 & 1234 & 2025 & Avg $\pm$ Std
 & 42 & 1234 & 2025 & Avg $\pm$ Std \\ \midrule

Mean Acc.
& 0.32 & 0.34 & 0.34 & 0.33 $\pm$ {\footnotesize 0.01}
& 0.41 & 0.37 & 0.41 & 0.40 $\pm$ {\footnotesize 0.02} \\

Self-Cert.
& 0.33 & 0.33 & 0.33 & 0.33 $\pm$ {\footnotesize 0.00}
& 0.37 & 0.33 & 0.33 & 0.34 $\pm$ {\footnotesize 0.02} \\

High Conf.
& 0.37 & 0.37 & 0.33 & 0.36 $\pm$ {\footnotesize 0.02}
& 0.40 & 0.33 & 0.33 & 0.36 $\pm$ {\footnotesize 0.03} \\

Low Ent.
& 0.33 & 0.40 & 0.30 & 0.34 $\pm$ {\footnotesize 0.04}
& 0.43 & 0.30 & 0.37 & 0.37 $\pm$ {\footnotesize 0.06} \\

\midrule
UID Metrics ($\downarrow$) \\
\midrule

Loc. non-uni (2$\sigma$)
& 0.33 & 0.33 & 0.40 & 0.36 $\pm$ {\footnotesize 0.03}
& 0.40 & 0.37 & 0.37 & 0.38 $\pm$ {\footnotesize 0.02} \\

\rowcolor{lightblue}
Loc. uni (2$\sigma$)
& 0.37 & 0.37 & 0.37 & 0.37 $\pm$ {\footnotesize 0.00}
& 0.37 & 0.33 & 0.40 & 0.37 $\pm$ {\footnotesize 0.03} \\

Loc. non-uni (3$\sigma$)
& 0.30 & 0.40 & 0.37 & 0.36 $\pm$ {\footnotesize 0.04}
& 0.40 & 0.33 & 0.33 & 0.36 $\pm$ {\footnotesize 0.03} \\

\rowcolor{lightblue}
Loc. uni (3$\sigma$)
& 0.33 & 0.40 & 0.43 & 0.39 $\pm$ {\footnotesize 0.04}
& 0.50 & 0.37 & 0.47 & 0.44 $\pm$ {\footnotesize 0.06} \\

\rowcolor{lightblue}
Glob. non-uni
& 0.40 & 0.33 & 0.37 & 0.37 $\pm$ {\footnotesize 0.03}
& 0.40 & 0.30 & 0.37 & 0.36 $\pm$ {\footnotesize 0.04} \\

Glob. uni
& 0.10 & 0.33 & 0.37 & 0.27 $\pm$ {\footnotesize 0.12}
& 0.47 & 0.33 & 0.40 & 0.40 $\pm$ {\footnotesize 0.06} \\

\bottomrule
\end{tabular}
}

\vspace{6pt}

\resizebox{0.98\textwidth}{!}{%
\begin{tabular}{@{}l|cccc|cccc@{}}
\toprule
\multirow{2}{*}{Methods ($\downarrow$)} &
\multicolumn{4}{c|}{HMMT2025} &
\multicolumn{4}{c}{MinervaMath} \\ \cmidrule(lr){2-5}\cmidrule(lr){6-9}
 & 42 & 1234 & 2025 & Avg $\pm$ Std
 & 42 & 1234 & 2025 & Avg $\pm$ Std \\ \midrule

Mean Acc.
& 0.20 & 0.19 & 0.22 & 0.20 $\pm$ {\footnotesize 0.01}
& 0.23 & 0.23 & 0.23 & 0.23 $\pm$ {\footnotesize 0.00} \\

Self-Cert.
& 0.20 & 0.17 & 0.20 & 0.19 $\pm$ {\footnotesize 0.02}
& 0.22 & 0.22 & 0.23 & 0.22 $\pm$ {\footnotesize 0.01} \\

High Conf.
& 0.23 & 0.20 & 0.20 & 0.21 $\pm$ {\footnotesize 0.02}
& 0.23 & 0.23 & 0.22 & 0.23 $\pm$ {\footnotesize 0.01} \\

Low Ent.
& 0.20 & 0.20 & 0.17 & 0.19 $\pm$ {\footnotesize 0.02}
& 0.21 & 0.23 & 0.22 & 0.22 $\pm$ {\footnotesize 0.01} \\

\midrule
UID Metrics ($\downarrow$) \\
\midrule

Loc. non-uni (2$\sigma$)
& 0.23 & 0.23 & 0.23 & 0.23 $\pm$ {\footnotesize 0.00}
& 0.22 & 0.25 & 0.21 & 0.23 $\pm$ {\footnotesize 0.02} \\

\rowcolor{lightblue}
Loc. uni (2$\sigma$)
& 0.17 & 0.20 & 0.17 & 0.18 $\pm$ {\footnotesize 0.02}
& 0.22 & 0.22 & 0.22 & 0.22 $\pm$ {\footnotesize 0.00} \\

Loc. non-uni (3$\sigma$)
& 0.17 & 0.20 & 0.23 & 0.20 $\pm$ {\footnotesize 0.03}
& 0.24 & 0.21 & 0.24 & 0.23 $\pm$ {\footnotesize 0.01} \\

\rowcolor{lightblue}
Loc. uni (3$\sigma$)
& 0.27 & 0.20 & 0.27 & 0.24 $\pm$ {\footnotesize 0.04}
& 0.24 & 0.22 & 0.24 & 0.23 $\pm$ {\footnotesize 0.01} \\

\rowcolor{lightblue}
Glob. non-uni
& 0.13 & 0.20 & 0.20 & 0.18 $\pm$ {\footnotesize 0.04}
& 0.21 & 0.21 & 0.21 & 0.21 $\pm$ {\footnotesize 0.00} \\

Glob. uni
& 0.30 & 0.27 & 0.23 & 0.27 $\pm$ {\footnotesize 0.03}
& 0.24 & 0.24 & 0.22 & 0.23 $\pm$ {\footnotesize 0.01} \\

\bottomrule
\end{tabular}
}

\end{table*}

%% file: Tables/ScaleModelSeed.tex
\newcommand{\pmfs}[1]{\,$\pm$\text{\footnotesize #1}}

\begin{table*}[t]
\caption{\textbf{Model Size Analysis.} Results of \texttt{Qwen3} models (1.7B, 4B, 8B) on AIME2025. Performance is reported across seeds 42, 1234, and 2025, with averages and standard deviations (Avg~$\pm$~Std).}
\label{tab:scalemodelseed}
\centering
\resizebox{1.0\textwidth}{!}{%
\begin{tabular}{@{}l|cccc|cccc|cccc@{}}
\toprule
\multirow{2}{*}{Methods ($\downarrow$)} &
\multicolumn{4}{c|}{1.7B} &
\multicolumn{4}{c|}{4B} &
\multicolumn{4}{c}{8B} \\ \cmidrule(lr){2-5}\cmidrule(lr){6-9}\cmidrule(lr){10-13}
 & {42} & {1234} & {2025} & {Avg $\pm$ Std}
 & {42} & {1234} & {2025} & {Avg $\pm$ Std}
 & {42} & {1234} & {2025} & {Avg $\pm$ Std} \\ \midrule
Mean Acc. 
& 0.33 & 0.34 & 0.37 & 0.35\pmfs{0.02}
& 0.66 & 0.65 & 0.64 & 0.65\pmfs{0.01}
& 0.66 & 0.66 & 0.68 & 0.67\pmfs{0.01} \\
Self-Cert. 
& 0.40 & 0.47 & 0.47 & 0.45\pmfs{0.04}
& 0.77 & 0.70 & 0.73 & 0.73\pmfs{0.03}
& 0.63 & 0.63 & 0.63 & 0.63\pmfs{0.00} \\
High Conf. 
& 0.33 & 0.33 & 0.43 & 0.37\pmfs{0.06}
& 0.67 & 0.57 & 0.63 & 0.62\pmfs{0.05}
& 0.60 & 0.60 & 0.60 & 0.60\pmfs{0.00} \\
Low Ent. 
& 0.33 & 0.33 & 0.43 & 0.37\pmfs{0.06}
& 0.73 & 0.57 & 0.60 & 0.63\pmfs{0.09}
& 0.60 & 0.60 & 0.60 & 0.60\pmfs{0.00} \\
\midrule
UID Metrics ($\downarrow$) \\
\midrule
Loc. non-uni (2$\sigma$) 
& 0.20 & 0.30 & 0.23 & 0.24\pmfs{0.05}
& 0.53 & 0.57 & 0.53 & 0.54\pmfs{0.02}
& 0.63 & 0.63 & 0.63 & 0.63\pmfs{0.00} \\
\rowcolor{lightblue}
Loc. uni (2$\sigma$) 
& 0.43 & 0.43 & 0.37 & 0.41\pmfs{0.04}
& 0.70 & 0.73 & 0.70 & 0.71\pmfs{0.02}
& 0.67 & 0.67 & 0.73 & 0.69\pmfs{0.04} \\
Loc. non-uni (3$\sigma$) 
& 0.20 & 0.30 & 0.23 & 0.24\pmfs{0.05}
& 0.53 & 0.57 & 0.53 & 0.54\pmfs{0.02}
& 0.63 & 0.63 & 0.63 & 0.63\pmfs{0.00} \\
\rowcolor{lightblue}
Loc. uni (3$\sigma$) 
& 0.43 & 0.43 & 0.37 & 0.41\pmfs{0.04}
& 0.70 & 0.70 & 0.67 & 0.69\pmfs{0.02}
& 0.67 & 0.67 & 0.73 & 0.69\pmfs{0.04} \\
\rowcolor{lightblue}
Glob. non-uni 
& 0.37 & 0.30 & 0.43 & 0.37\pmfs{0.07}
& 0.70 & 0.63 & 0.63 & 0.66\pmfs{0.04}
& 0.70 & 0.70 & 0.70 & 0.70\pmfs{0.00} \\
Glob. uni 
& 0.30 & 0.37 & 0.33 & 0.33\pmfs{0.03}
& 0.60 & 0.73 & 0.67 & 0.67\pmfs{0.07}
& 0.67 & 0.67 & 0.63 & 0.66\pmfs{0.02} \\
\bottomrule
\end{tabular}%
}
\end{table*}

%% file: Tables/ScaleSampleSeed.tex
\begin{table*}[t]
\caption{\textbf{Sample Size Analysis.} Results across different sampling strategies (\texttt{Sample by 3}, \texttt{5}, and \texttt{10}) on \texttt{AIME2025}. Performance is reported across seeds 42, 1234, and 2025, with averages and standard deviations (Avg~$\pm$~Std).}
\label{tab:sample-scale}
\centering
\resizebox{1.0\textwidth}{!}{%
\begin{tabular}{@{}l|cccc|cccc|cccc@{}}
\toprule
\multirow{2}{*}{Methods ($\downarrow$)} &
\multicolumn{4}{c|}{Sample by 3} &
\multicolumn{4}{c|}{Sample by 5} &
\multicolumn{4}{c}{Sample by 10} \\ 
\cmidrule(lr){2-5}\cmidrule(lr){6-9}\cmidrule(lr){10-13}
 & {42} & {1234} & {2025} & {Avg $\pm$ Std}
 & {42} & {1234} & {2025} & {Avg $\pm$ Std}
 & {42} & {1234} & {2025} & {Avg $\pm$ Std} \\ 
\midrule
Mean Acc.   
& 0.67 & 0.67 & 0.67 & 0.67{\footnotesize$\pm$0.00}
& 0.66 & 0.66 & 0.68 & 0.67{\footnotesize$\pm$0.01}
& 0.68 & 0.67 & 0.69 & 0.68{\footnotesize$\pm$0.01} \\
Self-Cert.  
& 0.70 & 0.70 & 0.70 & 0.70{\footnotesize$\pm$0.00}
& 0.63 & 0.63 & 0.63 & 0.63{\footnotesize$\pm$0.00}
& 0.57 & 0.63 & 0.67 & 0.62{\footnotesize$\pm$0.04} \\
High Conf.  
& 0.63 & 0.63 & 0.63 & 0.63{\footnotesize$\pm$0.00}
& 0.60 & 0.60 & 0.60 & 0.60{\footnotesize$\pm$0.00}
& 0.57 & 0.53 & 0.60 & 0.57{\footnotesize$\pm$0.03} \\
Low Ent.    
& 0.63 & 0.63 & 0.63 & 0.63{\footnotesize$\pm$0.00}
& 0.60 & 0.60 & 0.60 & 0.60{\footnotesize$\pm$0.00}
& 0.50 & 0.57 & 0.60 & 0.56{\footnotesize$\pm$0.04} \\
\midrule
UID Metrics ($\downarrow$) \\
\midrule
Loc. non-uni (2$\sigma$) 
& 0.63 & 0.63 & 0.63 & 0.63{\footnotesize$\pm$0.00}
& 0.63 & 0.63 & 0.63 & 0.63{\footnotesize$\pm$0.00}
& 0.50 & 0.50 & 0.60 & 0.53{\footnotesize$\pm$0.05} \\
\rowcolor{lightblue}
Loc. uni (2$\sigma$)     
& 0.73 & 0.73 & 0.73 & 0.73{\footnotesize$\pm$0.00}
& 0.67 & 0.67 & 0.73 & 0.69{\footnotesize$\pm$0.03}
& 0.70 & 0.73 & 0.73 & 0.72{\footnotesize$\pm$0.02} \\
Loc. non-uni (3$\sigma$)
& 0.63 & 0.63 & 0.63 & 0.63{\footnotesize$\pm$0.00}
& 0.63 & 0.63 & 0.63 & 0.63{\footnotesize$\pm$0.00}
& 0.50 & 0.50 & 0.60 & 0.53{\footnotesize$\pm$0.05} \\
\rowcolor{lightblue}
Loc. uni (3$\sigma$)     
& 0.73 & 0.73 & 0.73 & 0.73{\footnotesize$\pm$0.00}
& 0.67 & 0.67 & 0.73 & 0.69{\footnotesize$\pm$0.03}
& 0.70 & 0.73 & 0.73 & 0.72{\footnotesize$\pm$0.02} \\
\rowcolor{lightblue}
Glob. non-uni            
& 0.70 & 0.70 & 0.70 & 0.70{\footnotesize$\pm$0.00}
& 0.70 & 0.70 & 0.70 & 0.70{\footnotesize$\pm$0.00}
& 0.70 & 0.67 & 0.73 & 0.70{\footnotesize$\pm$0.03} \\
Glob. uni                
& 0.70 & 0.70 & 0.70 & 0.70{\footnotesize$\pm$0.00}
& 0.67 & 0.67 & 0.63 & 0.66{\footnotesize$\pm$0.02}
& 0.63 & 0.63 & 0.63 & 0.63{\footnotesize$\pm$0.00} \\
\bottomrule
\end{tabular}%
}
\end{table*}

%% file: Tables/MajorityVoting.tex
\begin{table*}[t]
\caption{\textbf{UID-based selection with response-level aggregation.}
We use Borda Voting (majority voting under matched sampling budgets) adopted from~\citet{kang2025scalablebestofnselectionlarge} on AIME 2025.}
\centering
\setlength{\tabcolsep}{10pt}
\resizebox{\linewidth}{!}{%
\begin{tabular}{l|c|c|c}
\toprule
Method ($\downarrow$)
& \texttt{Qwen3-8B}
& \texttt{DS-R1-Distill-Qwen-7B}
& \texttt{DS-R1-Distill-Llama-8B} \\
\midrule
Majority Voting
& 0.67
& 0.43
& 0.27 \\

\midrule
UID Metrics ($\downarrow$) \\
\midrule

\rowcolor{lightblue}
Loc.\ uni
& \textbf{0.73}
& 0.43
& \textbf{0.40} \\

\rowcolor{lightblue}
Loc uni + Borda
& \textbf{0.73}
& 0.43
& 0.37 \\

\rowcolor{lightblue}
Glob.\ non-uni
& 0.70
& \textbf{0.47}
& 0.33 \\

\rowcolor{lightblue}
Glob. non-uni + Borda
& 0.70
& \textbf{0.47}
& 0.33 \\


\bottomrule
\end{tabular}
}
\vspace{-0.1in}
\label{tab:majority_voting}
\end{table*}

%% file: Tables/Prompts.tex
\begin{table*}[ht]
\centering
\renewcommand{\arraystretch}{1.2}
\begin{tabular}{p{3cm} p{12cm}}
\toprule
\textbf{Settings} &
\textbf{Content} \\ \midrule

Communication Prompting &
\texttt{Please answer the following math question as if you are explaining it to a listener.} \\
& \texttt{Communicate your reasoning clearly and naturally.} \\
& \texttt{You should provide your final answer in the format \textbackslash boxed\{YOUR\_ANSWER\}.} \\
& \texttt{\textbackslash n\textbackslash n Question:\textbackslash n\{question\}\textbackslash n} \\ \midrule

Naive Prompting &
\texttt{Please answer the following math question.} \\
& \texttt{You should provide your final answer in the format \textbackslash boxed\{YOUR\_ANSWER\}.} \\
& \texttt{\textbackslash n\textbackslash n Question:\textbackslash n\{question\}\textbackslash n} \\

\bottomrule
\end{tabular}

\caption{Prompts used for Communicative vs. Naive Prompting.}
\label{tab:prompts}

\end{table*}

%% file: Tables/K_Sweep_Seed42.tex
\begin{table*}[t]
\caption{\textbf{$\tau$-sweep results ($\tau$ = 2–5).} Performance on AIME2025, BRUMO2025, HMMT2025, and MinervaMath (MM). Results are from seed 42.}
\centering
\resizebox{\textwidth}{!}{
\begin{tabular}{@{}l|cccc|cccc|cccc@{}}
\toprule
 & \multicolumn{4}{c|}{\texttt{DS-R1-Distill-Qwen-7B}} 
 & \multicolumn{4}{c|}{\texttt{DS-R1-Distill-Llama-8B}}
 & \multicolumn{4}{c}{\texttt{Qwen3-8B}} \\
\cmidrule(lr){2-5} \cmidrule(lr){6-9} \cmidrule(lr){10-13}
Methods ($\downarrow$) 
& AIME & BRUMO & HMMT & MM
& AIME & BRUMO & HMMT & MM
& AIME & BRUMO & HMMT & MM \\
\midrule

\midrule
\multicolumn{13}{l}{\textbf{$\tau$ = 2}} \\
Loc. non-uni 
& 0.23 & 0.40 & 0.30 & 0.27
& 0.33 & 0.47 & 0.17 & 0.26
& 0.70 & 0.70 & 0.47 & 0.27 \\
Loc. uni     
& 0.43 & 0.43 & 0.27 & 0.30
& 0.37 & 0.50 & 0.17 & 0.27
& 0.73 & 0.67 & 0.30 & 0.30 \\

\midrule
\multicolumn{13}{l}{\textbf{$\tau$ = 3}} \\
Loc. non-uni 
& 0.23 & 0.40 & 0.17 & 0.28
& 0.43 & 0.43 & 0.20 & 0.27
& 0.70 & 0.70 & 0.50 & 0.28 \\
Loc. uni     
& 0.40 & 0.47 & 0.27 & 0.30
& 0.33 & 0.53 & 0.13 & 0.28
& 0.70 & 0.70 & 0.37 & 0.30 \\

\midrule
\multicolumn{13}{l}{\textbf{$\tau$ = 4}} \\
Loc. non-uni 
& 0.33 & 0.50 & 0.20 & 0.29
& 0.43 & 0.43 & 0.23 & 0.28
& 0.67 & 0.73 & 0.47 & 0.29 \\
Loc. uni     
& 0.43 & 0.47 & 0.20 & 0.31
& 0.40 & 0.57 & 0.20 & 0.29
& 0.60 & 0.67 & 0.37 & 0.31 \\

\midrule
\multicolumn{13}{l}{\textbf{$\tau$ = 5}} \\
Loc. non-uni 
& 0.30 & 0.43 & 0.23 & 0.30
& 0.43 & 0.50 & 0.20 & 0.29
& 0.67 & 0.70 & 0.43 & 0.30 \\
Loc. uni     
& 0.40 & 0.50 & 0.20 & 0.32
& 0.37 & 0.53 & 0.17 & 0.29
& 0.63 & 0.67 & 0.33 & 0.32 \\

\bottomrule
\end{tabular}
}
\label{tab:ksweep_seed42}
\end{table*}